\documentclass[letterpaper, 10 pt, conference]{ieeeconf}

\usepackage{amsmath,amsthm,amssymb}
\usepackage{mathtools}
\usepackage{graphicx}
\usepackage{float}
\usepackage{subcaption}
\usepackage{bm}
\usepackage{tikz}
\usepackage{tikz-3dplot}
\usepackage{pgfplots}
\usepackage{pgfplotstable}
\usepackage{array}
\usepackage{multirow}
\usepackage{booktabs}
\let\labelindent\relax
\usepackage{enumitem}
\usepackage{makecell}
\usepackage{silence}
\usepackage{url}
\usepackage{todonotes}
\usepackage[nolist,nohyperlinks]{acronym}
\usepackage[ruled,linesnumbered]{algorithm2e}
\usepackage[hidelinks]{hyperref}
\usepackage[capitalize]{cleveref}
\usepackage[numbered,open]{bookmark}
\usepackage[notransparent]{svg}
\usepackage[normalem]{ulem}
\usepackage[export]{adjustbox}

\renewcommand{\vec}[1]{\mathbf{\boldsymbol{#1}}}
\newcommand{\mat}[1]{\boldsymbol{\mathbf{#1}}}

\newcommand{\minimize}{\mathop{\text{minimize}}}

\newcommand{\subjectto}{\mathop{\text{subject to}}}
\newcommand{\der}{\text{r}}
\newcommand{\defeq}{\triangleq}

\theoremstyle{definition}

\usetikzlibrary{arrows.meta}
\usetikzlibrary{3d}
\usetikzlibrary{calc}
\usetikzlibrary{positioning}
\usetikzlibrary{fit}
\usetikzlibrary{angles}
\usetikzlibrary{quotes}
\usetikzlibrary{patterns}
\usetikzlibrary{pgfplots.statistics}

\usepgfplotslibrary{colorbrewer}
\pgfplotsset{compat=1.9}
\pgfplotsset{legend cell align={left}}
\pgfplotsset{every axis plot/.append style={line width=0.8pt}}
\pgfplotsset{cycle list/Set1-5}

\newcolumntype{L}[1]{>{\raggedright\let\newline\\\arraybackslash\hspace{0pt}}m{#1}}
\newcolumntype{C}[1]{>{\centering\let\newline\\\arraybackslash\hspace{0pt}}m{#1}}
\newcolumntype{R}[1]{>{\raggedleft\let\newline\\\arraybackslash\hspace{0pt}}m{#1}}

\makeatletter
\newcommand{\nosemic}{\renewcommand{\@endalgocfline}{\relax}}
\newcommand{\dosemic}{\renewcommand{\@endalgocfline}{\algocf@endline}}
\makeatother
\SetAlgoCaptionSeparator{}

\DeclareRobustCommand{\svdots}{
  \vbox{%
    \baselineskip=0.33333\normalbaselineskip
    \lineskiplimit=0pt
    \hbox{.}\hbox{.}\hbox{.}%
    \kern-0.2\baselineskip
  }%
}

\newcommand{\glen}[1]{{\textcolor[rgb]{1,0,0}{\textsf{[Glen: {#1}]}}}}
\renewcommand{\glen}[1]{}

\IEEEoverridecommandlockouts

\begin{document}

\title{A Convex Formulation of Compliant Contact between Filaments and Rigid Bodies}

\author{Wei-Chen Li and Glen Chou
\thanks{The authors are with the Georgia Institute of Technology, Atlanta, GA, USA. \texttt{\{wli777, chou\}@gatech.edu}}
}

\maketitle

\begin{acronym}

\newacro{DER}{discrete elastic rod}
\newacro{FEM}{finite element method}
\newacro{PBD}{position-based dynamics}
\newacro{IPC}{incremental potential contact}
\newacro{KKT}{Karush-Kuhn-Tucker}
\newacro{NCP}{nonlinear complementarity problem}
\newacro{SAP}{semi-analytic primal}
\newacro{PD}{proportional–derivative}
\newacro{DoFs}{degrees of freedom}

\end{acronym}

\begin{abstract}
We present a computational framework for simulating filaments interacting with rigid bodies through contact. Filaments are challenging to simulate due to their codimensionality, i.e., they are one-dimensional structures embedded in three-dimensional space. Existing methods often assume that filaments remain permanently attached to rigid bodies. Our framework unifies \ac{DER} modeling, a pressure field patch contact model, and a convex contact formulation to accurately simulate frictional interactions between slender filaments and rigid bodies -- capabilities not previously achievable. Owing to the convex formulation of contact, each time step can be solved to global optimality, guaranteeing complementarity between contact velocity and impulse. We validate the framework by assessing the accuracy of frictional forces and comparing its physical fidelity against baseline methods. Finally, we demonstrate its applicability in both soft robotics, such as a stochastic filament-based gripper, and deformable object manipulation, such as shoelace tying, providing a versatile simulator for systems involving complex filament-filament and filament-rigid body interactions.
\end{abstract}

\section{Introduction}
The simulation of robotic systems has proven invaluable for controller design, robot learning, data generation, and system verification. Despite widespread reliance on simulation software, few toolkits can accurately model slender deformable objects such as ropes, cables, and shoelaces with high physical fidelity. These objects are central to many applications, including knot tying, cable routing, and surgical suturing, yet they undergo large deformations, self-contact, and topological changes such as knotting, while also making and breaking contact with external bodies -- factors that make them especially challenging to simulate. Existing approaches either employ simplified models, which may be physically inaccurate \cite{RN797}, or rely on full volumetric \ac{FEM}, which can be computationally prohibitive for the fine meshes required by slender objects, as the computational cost typically grows superlinearly with the number of mesh elements \cite{RN1537}.
This work addresses this gap by presenting a simulation framework for slender deformable objects that combines discrete elastic rod (DER) modeling, a pressure field patch contact model, and a convex contact formulation to enable accurate simulation of contact with rigid bodies.
Our efficient and accurate simulation framework aims to support both the characterization of soft robotics and the development of control strategies for deformable object manipulation \cite{RN815}.

\begin{figure}
    \centering
    \subcaptionbox{\label{fig:knot-tightening-a}}{
        \includegraphics[width=0.32\linewidth]{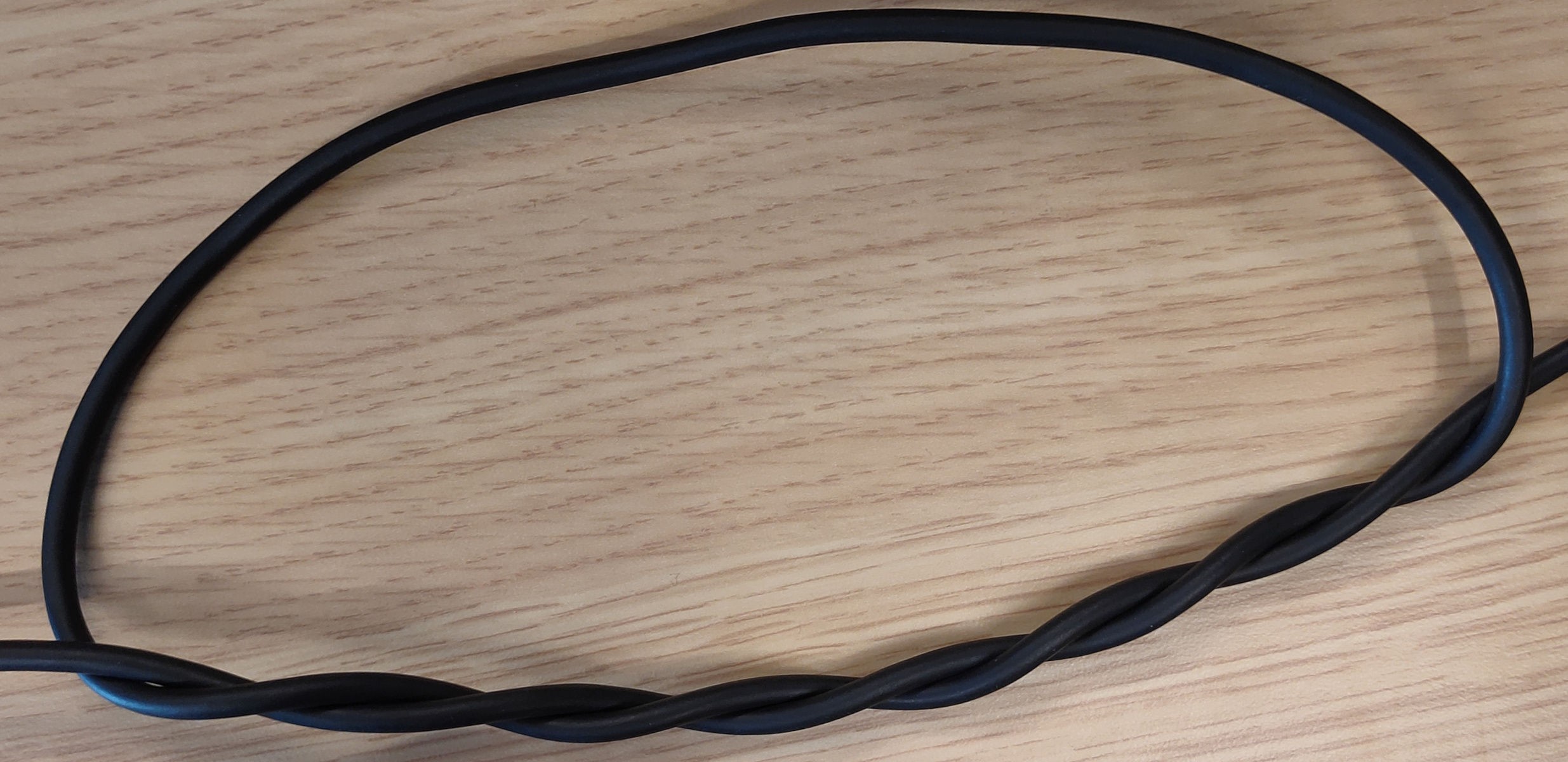}
        \includegraphics[width=0.32\linewidth]{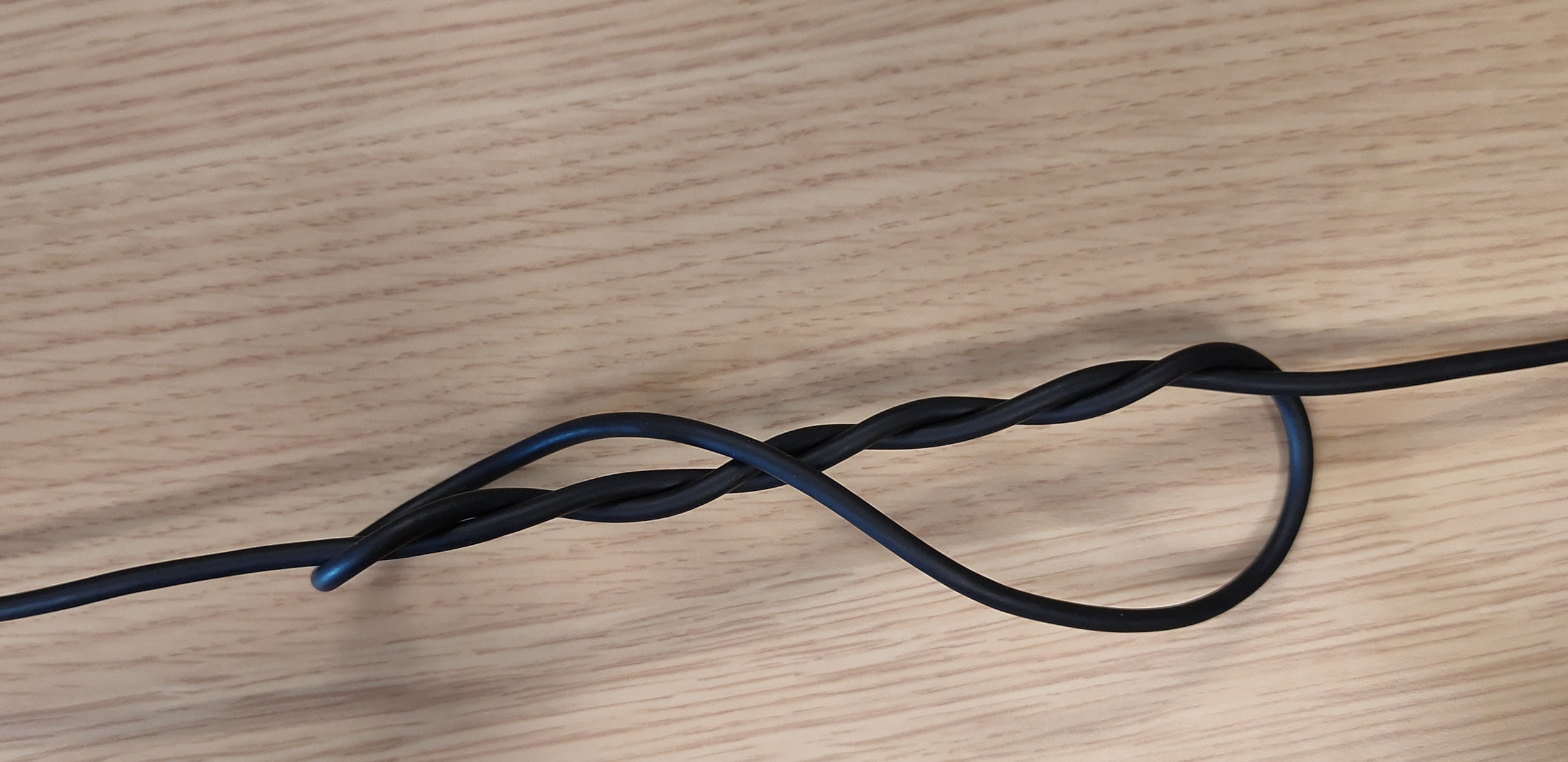}
        \includegraphics[width=0.32\linewidth]{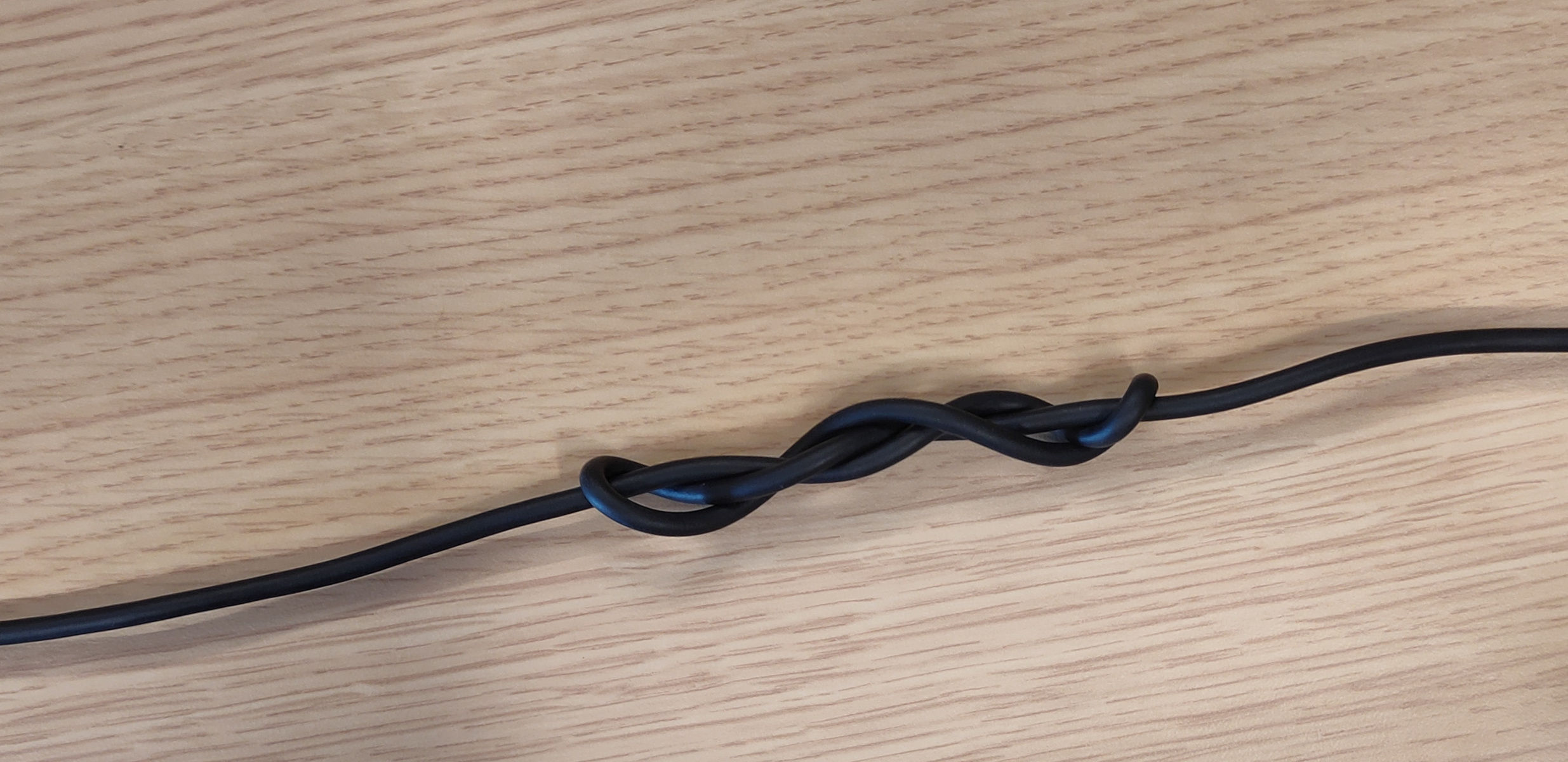}
    }
    \subcaptionbox{\label{fig:knot-tightening-b}}{
        \adjincludegraphics[width=0.32\linewidth, trim={{.19\width} {.20\height} {.22\width} {.19\height}}, clip]{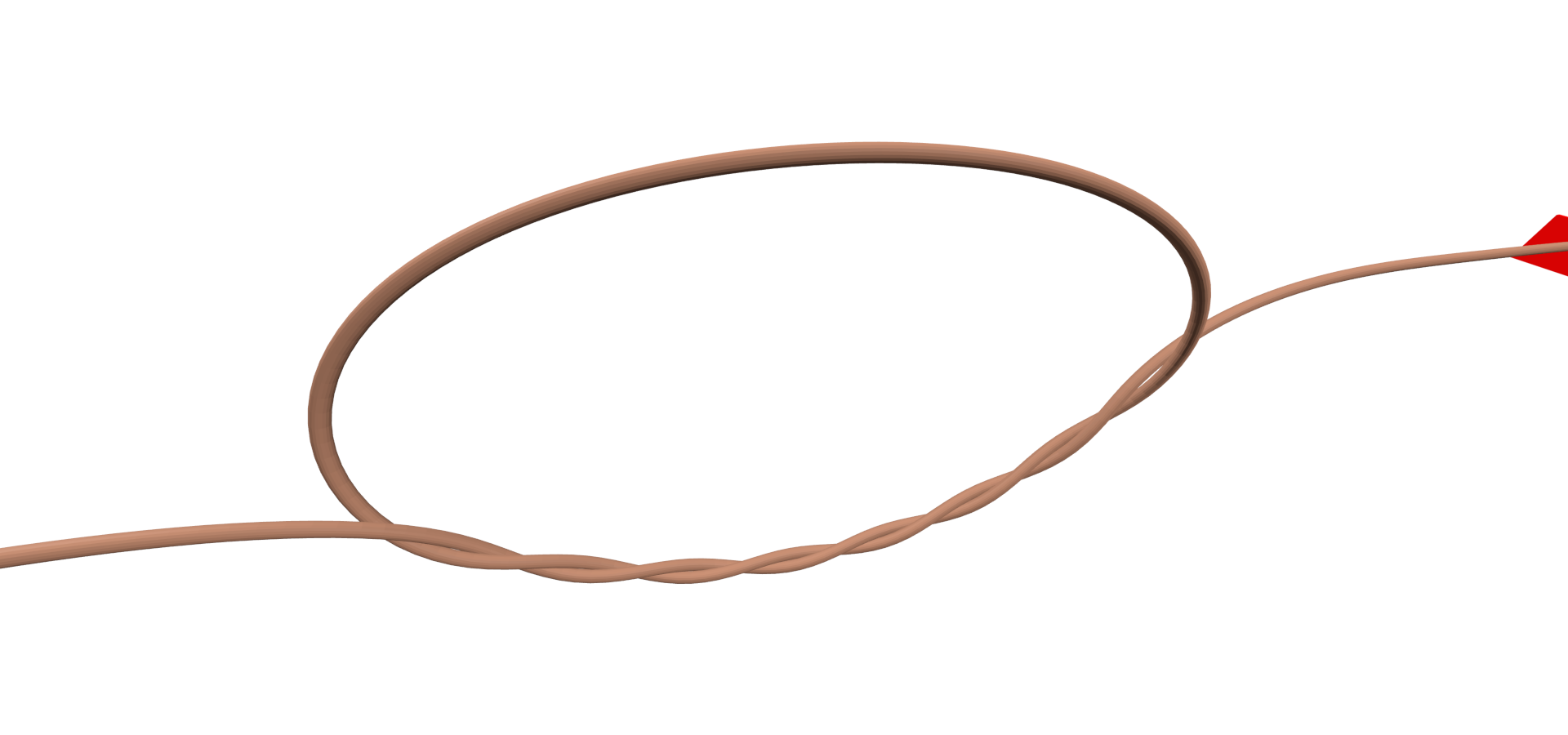}
        \adjincludegraphics[width=0.32\linewidth, trim={{.19\width} {.20\height} {.22\width} {.19\height}}, clip]{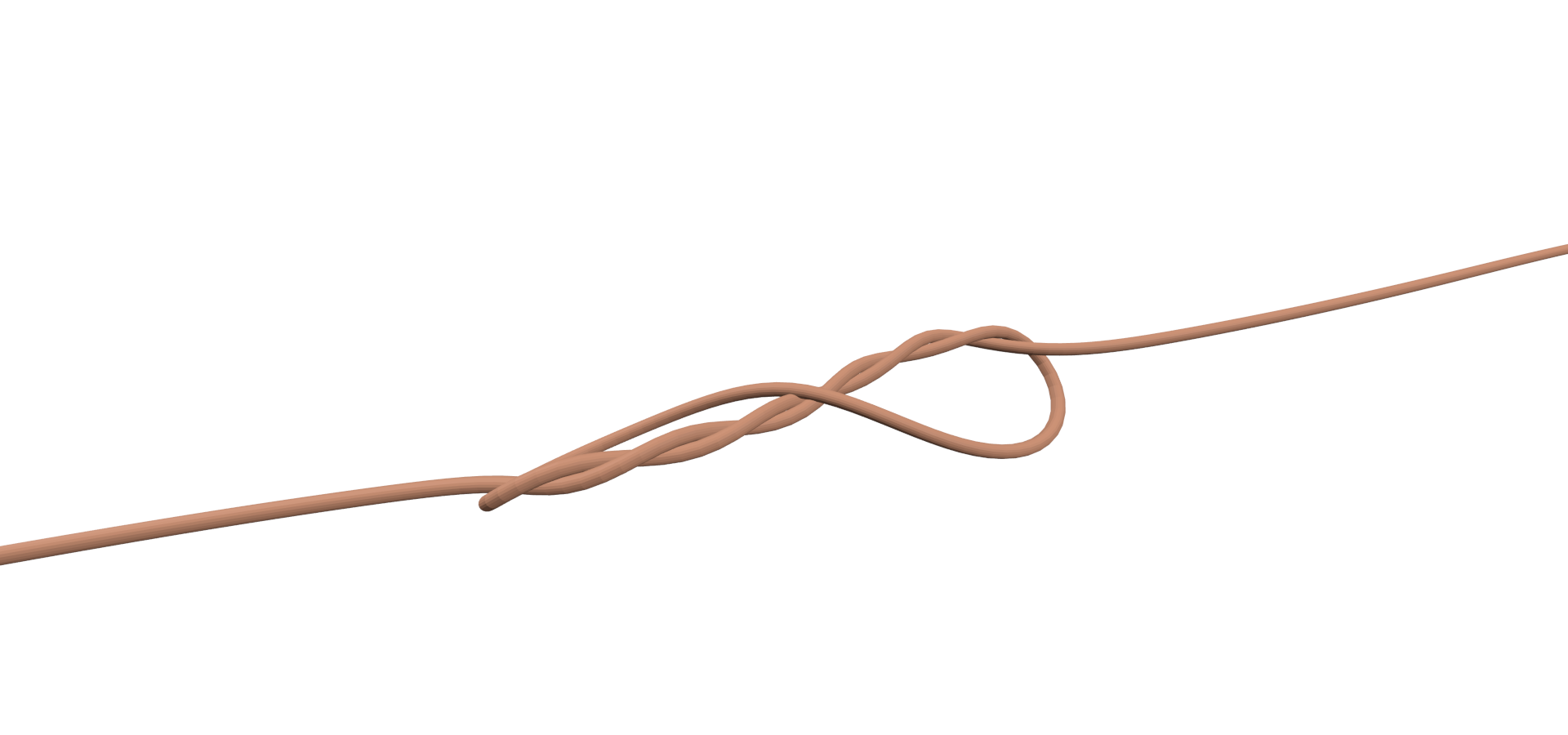}
        \adjincludegraphics[width=0.32\linewidth, trim={{.19\width} {.20\height} {.22\width} {.19\height}}, clip]{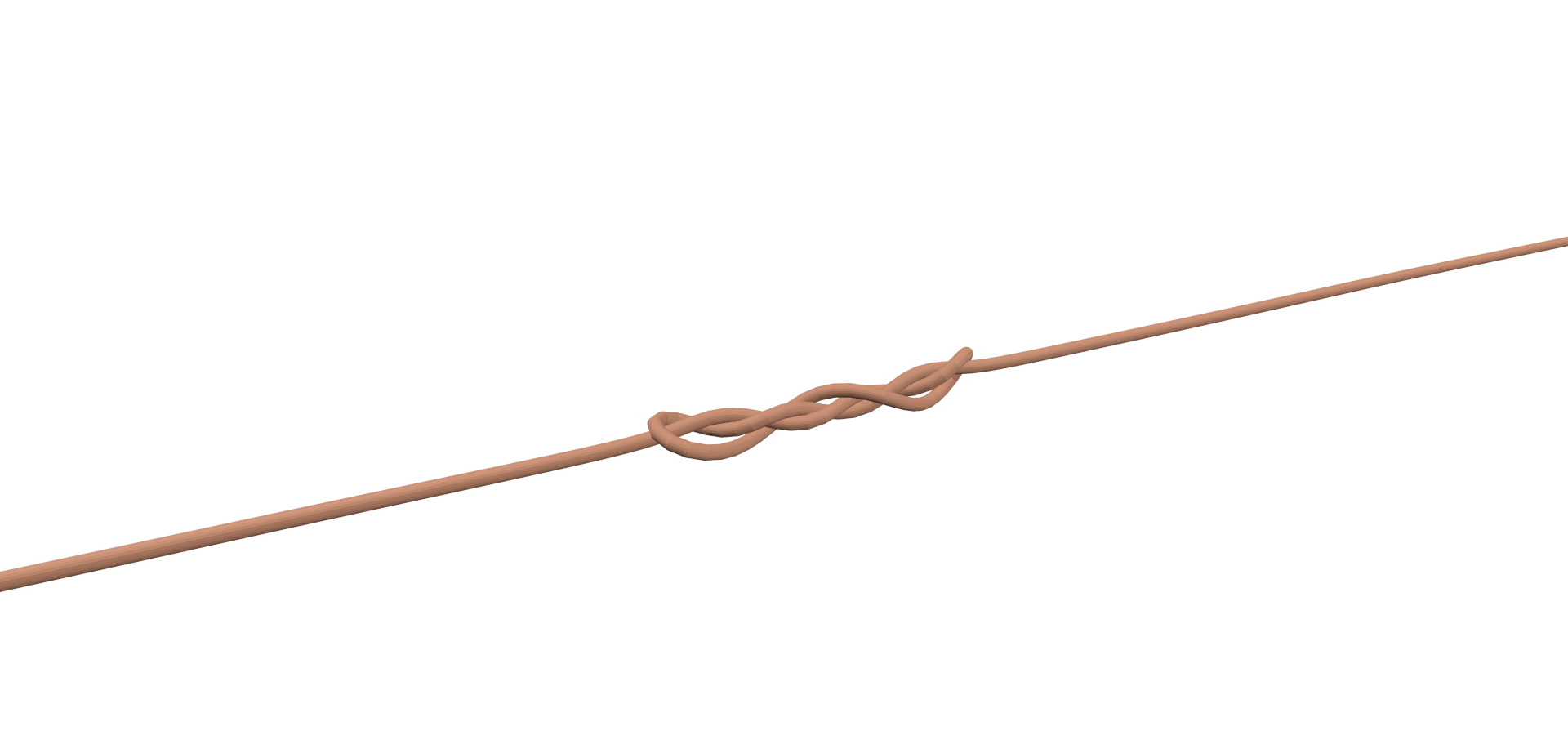}
    }
    \subcaptionbox{\label{fig:knot-tightening-c}}{
        \adjincludegraphics[width=0.32\linewidth, trim={{.19\width} {.14\height} {.20\width} {.12\height}}, clip]{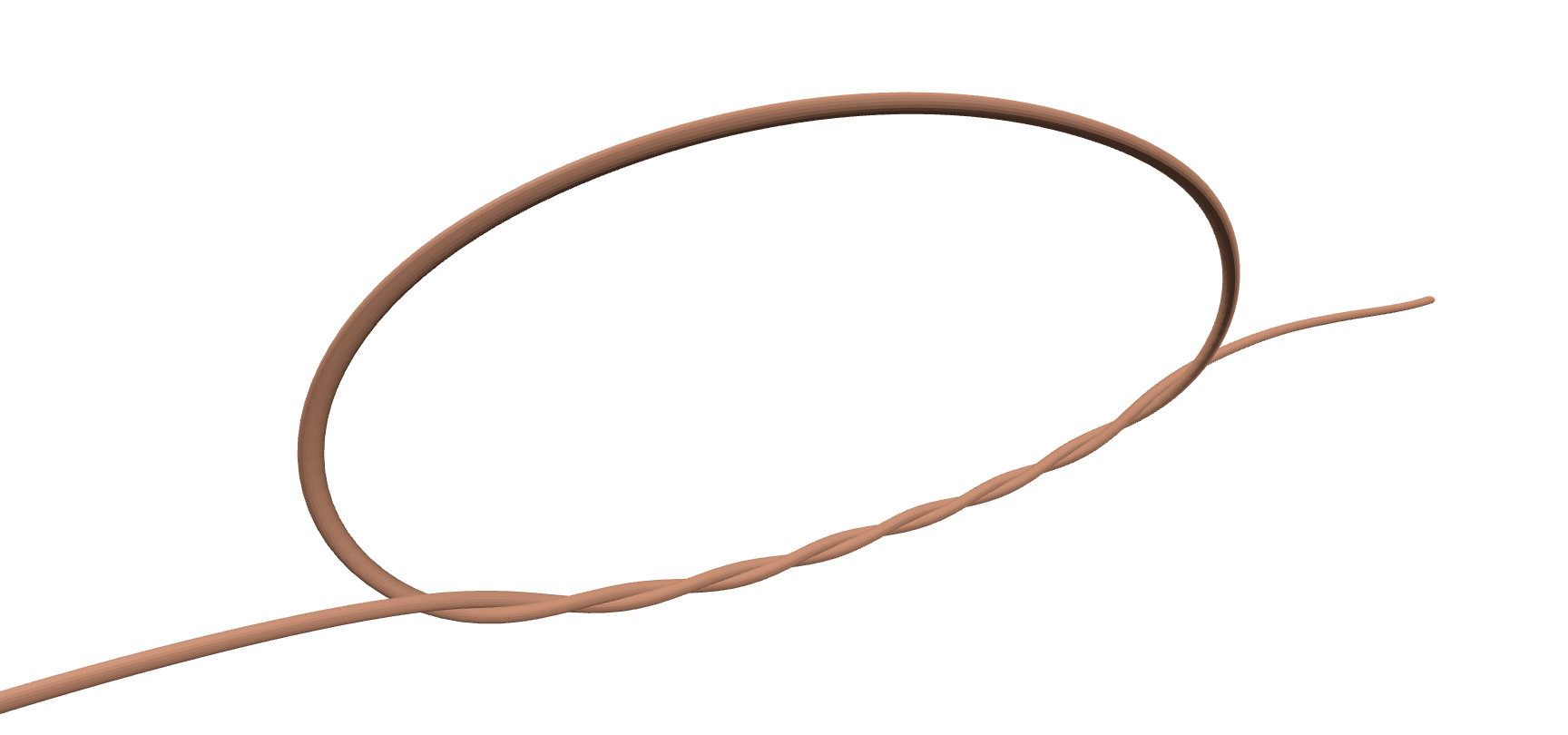}
        \adjincludegraphics[width=0.32\linewidth, trim={{.19\width} {.14\height} {.20\width} {.12\height}}, clip]{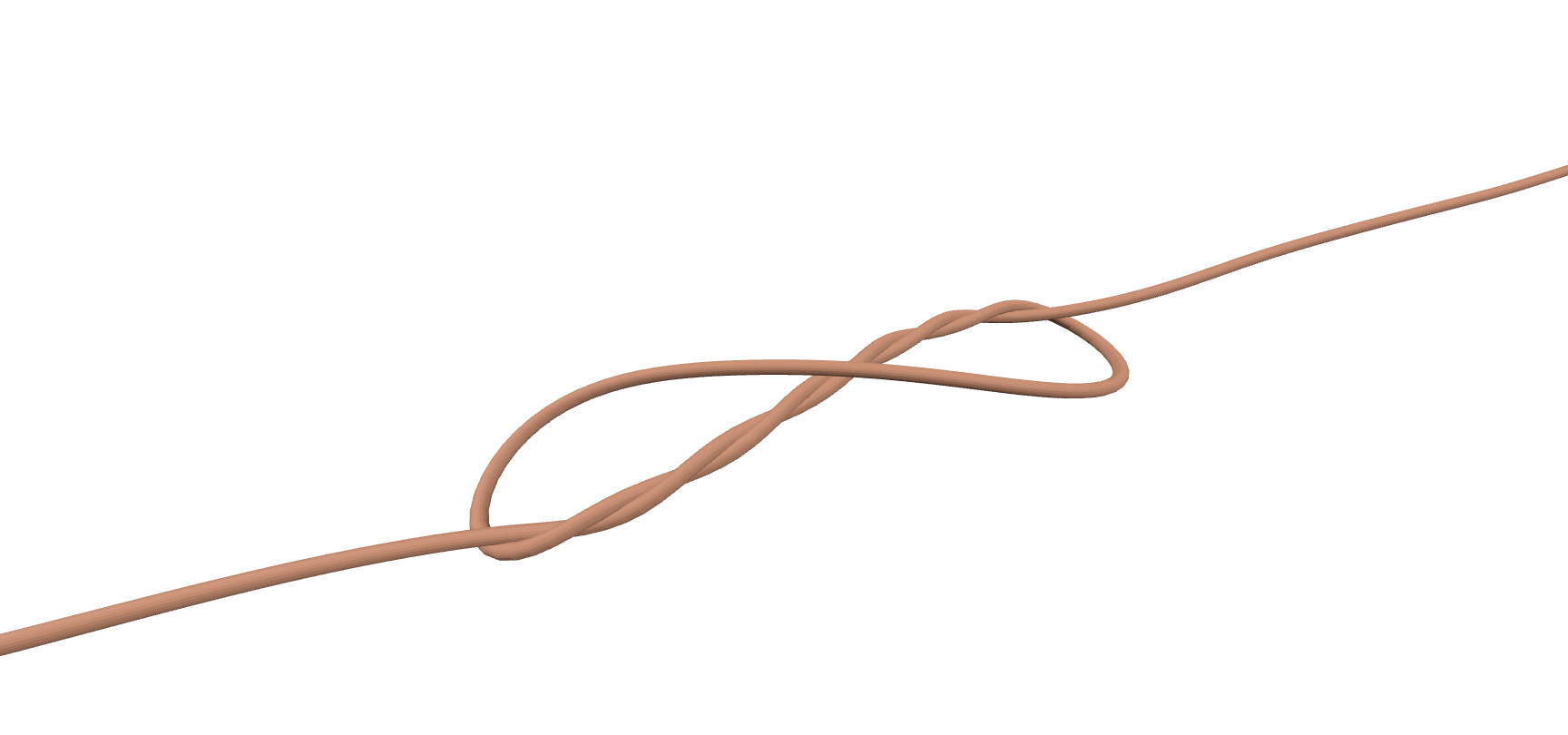}
        \adjincludegraphics[width=0.32\linewidth, trim={{.19\width} {.14\height} {.20\width} {.12\height}}, clip]{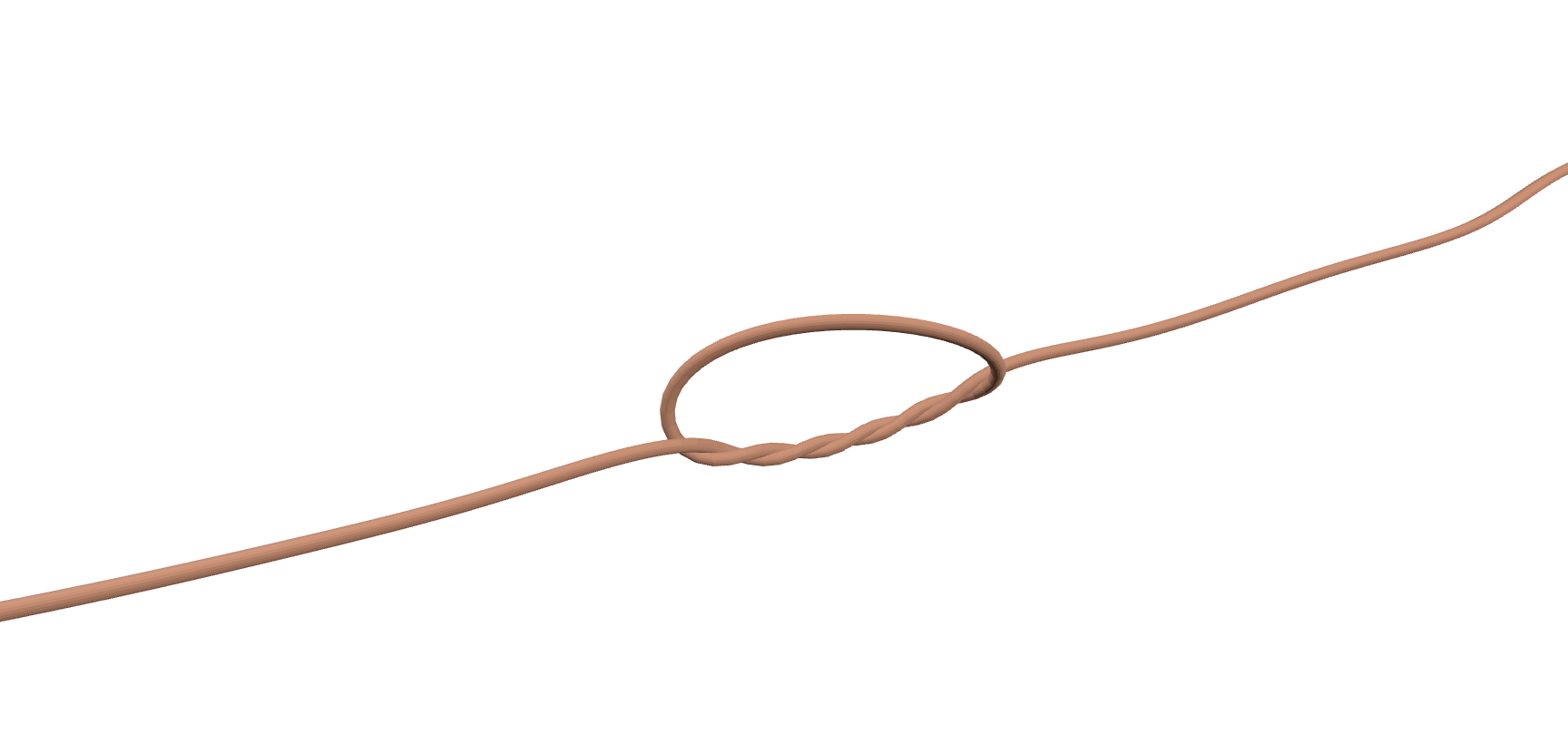}
    }
    \subcaptionbox{\label{fig:knot-tightening-d}}{
        \adjincludegraphics[width=0.32\linewidth, trim={{.22\width} {.13\height} {.22\width} {.13\height}}, clip]{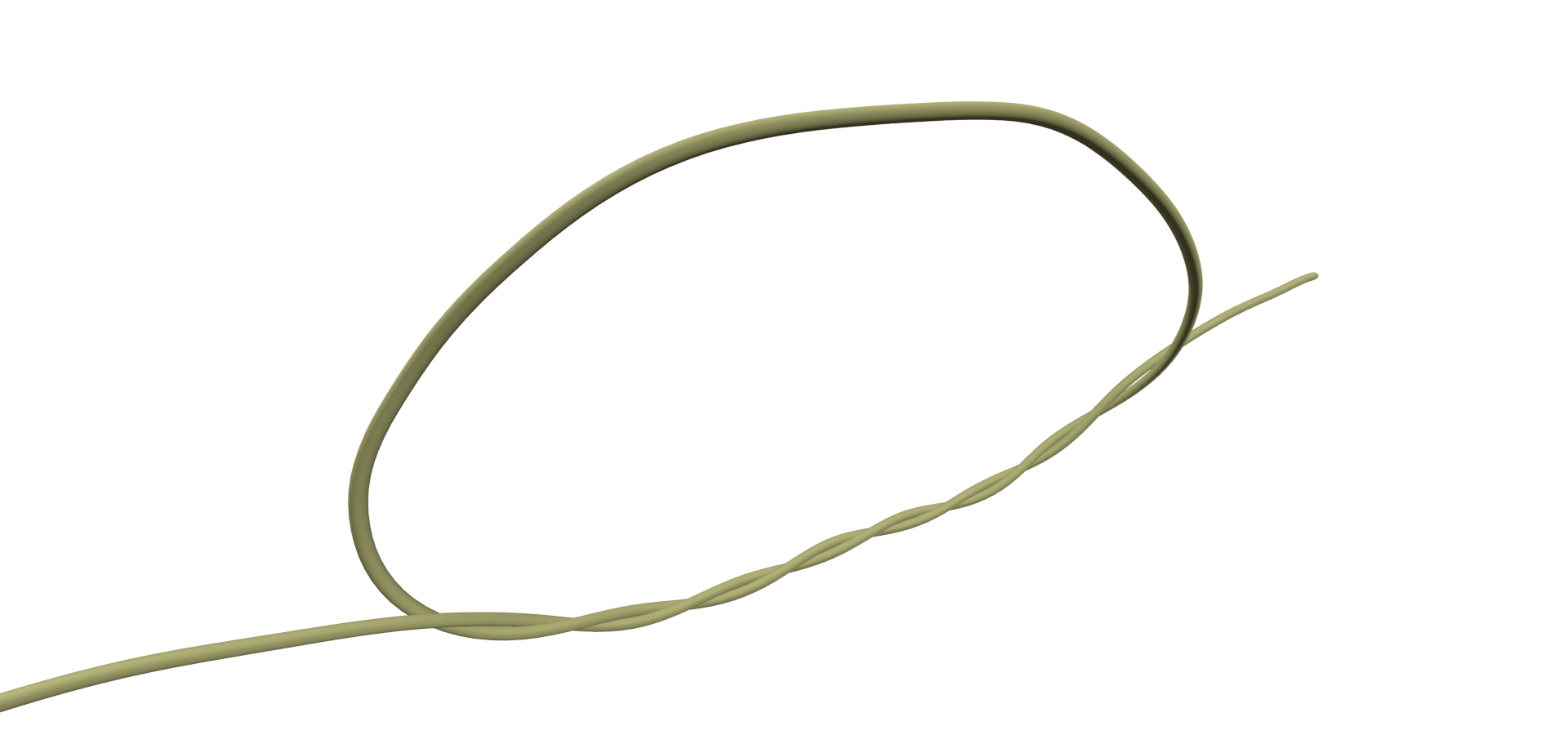}
        \adjincludegraphics[width=0.32\linewidth, trim={{.22\width} {.13\height} {.22\width} {.13\height}}, clip]{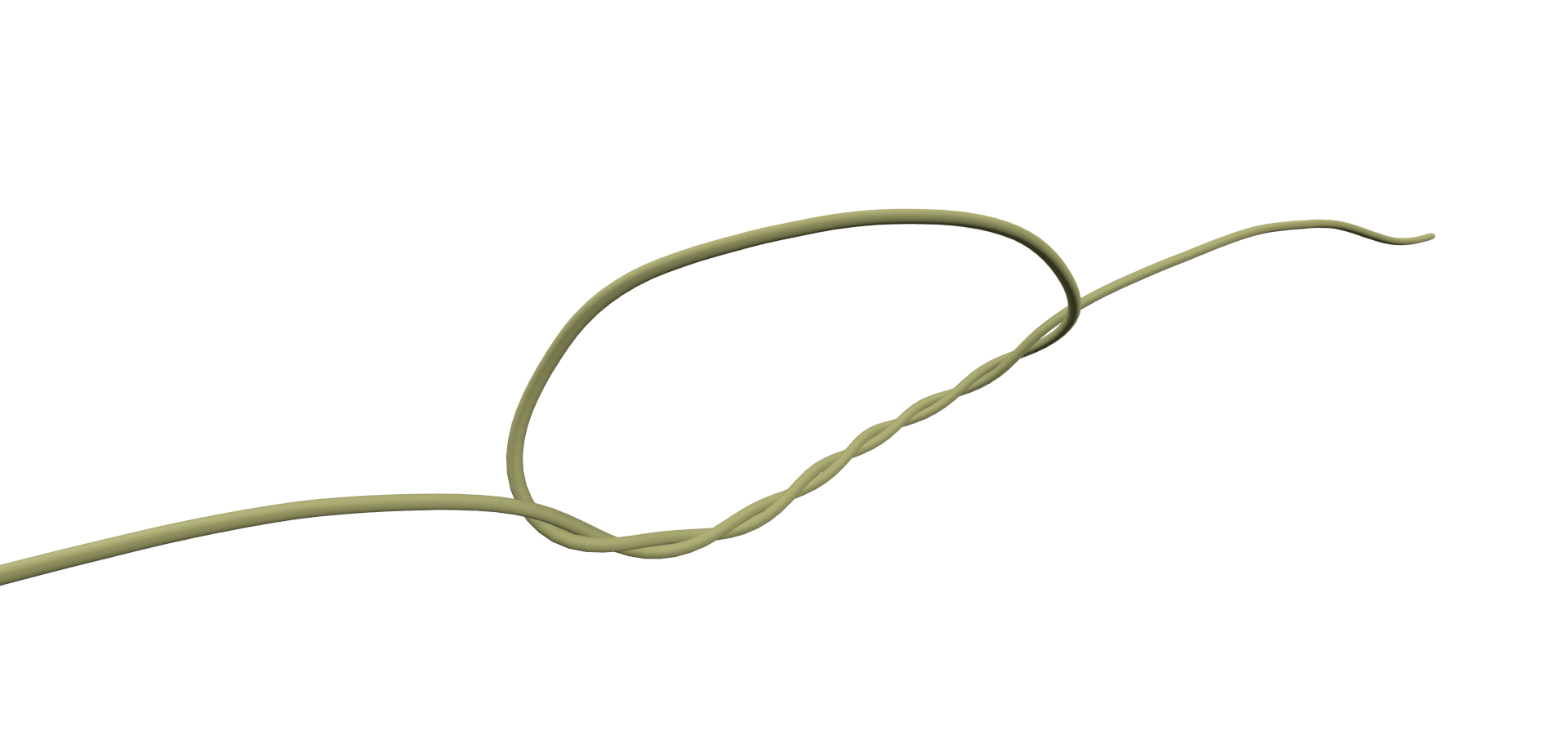}
        \adjincludegraphics[width=0.32\linewidth, trim={{.22\width} {.13\height} {.22\width} {.13\height}}, clip]{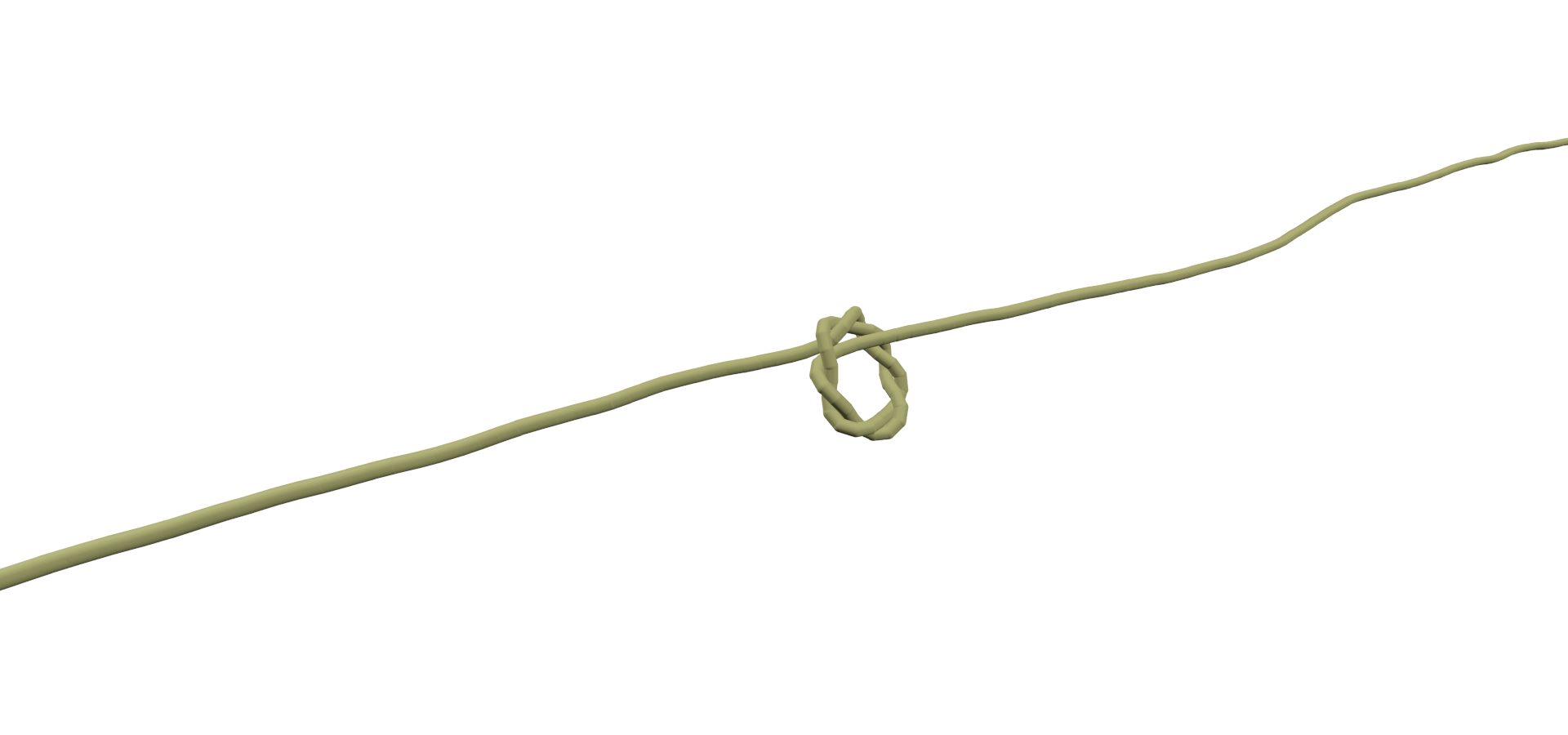}
    }
    \caption{Keyframes of knot tightening with an unknotting number of four, achieved by pulling on both ends of a rope.
        (a) Real rope.
        (b) Our simulation framework.
        (c) Existing contact energy-based method \cite{RN773}, which fails to prevent interpenetration; in the final keyframe, the rope undergoes tunneling, reducing the unknotting number to three.
        (d) A chain of rigid capsules with bending and twisting springs, which fails to reproduce realistic knot tightening behavior.}
    \label{fig:knot-tightening}
\end{figure}

We implement our framework on top of Drake \cite{drake}. Our main contributions are as follows:
\begin{enumerate}[leftmargin=*]
    \item \textit{Convex contact formulation for filaments.} We develop a convex contact solver, previously limited to rigid and volumetric deformables, that can simulate slender deformable bodies.
    \item \textit{Patch contact for slender deformables.} We extend a pressure field patch contact model, previously applied to rigid bodies, for slender deformables. This enables the simulation of lateral compliance and distributed contact forces that go beyond traditional point contact models.
    \item \textit{Demonstration in robotics applications.} We validate the framework on physics benchmarks such as knot tightening and showcase novel robotic applications including shoelace tying, which was not feasible with previous point contact models due to the highly variable pinching forces, preventing stable friction forces from being established and allowing the shoelace to slip.
\end{enumerate}
Our source code is available at \url{https://github.com/wei-chen-li/drake-DER}.

\section{Related Work}
The accurate simulation of filaments requires both a faithful mechanical model of slender deformables, and a robust treatment of self-contact and contact with external bodies. Prior research has approached these two aspects largely independently: studies on filament mechanics have focused on capturing bending, twisting, and stretching behaviors, while work on contact modeling has concentrated on non-penetration enforcement and frictional interactions. In the following, we review existing approaches in these two areas.

\subsection{Models for Filaments}
Rigid body simulators can be adapted to model filaments by connecting a series of rigid capsules via spherical joints, with springs used to approximate the twisting and bending stiffness of the filament. Previous work has employed simulators such as MuJoCo and PyBullet for this purpose \cite{RN798, lim2022real2sim2real}. While this approach is straightforward, it requires careful tuning of the spring stiffness parameters and does not guarantee convergence to the physically correct behavior as the model resolution is increased.

\Ac{PBD} models materials as a set of particles with holonomic constraints. Future particle positions are first predicted from the current positions and velocities, and then projected onto the constraint manifold to enforce the constraints \cite{muller2007position}. PhysX employs this approach for simulating cloth and rope, where ropes are represented using distance and angular constraints, as well as additional constraints for interactions with external objects. A major limitation of this method is that the effective stiffness depends on the integration time step, which limits its ability to accurately simulate force effects.

The full mechanics of filaments can be described by the Cosserat rod model \cite{RN762}, which governs the dynamics of slender bodies capable of bending, twisting, stretching, and shearing. Because shear deformation is explicitly modeled, each rod segment is associated with an $\mathrm{SO}(3)$ frame. Elastica \cite{RN819} employs this model for simulating biological muscles, where shear deformations are essential. However, the use of explicit Verlet integration restricts the simulator to small time steps, limiting simulation speed.

Neglecting shear deformation enables the use of implicit integrators, allowing simulations to advance with larger time steps. This approach corresponds to the \ac{DER} model \cite{RN802}. Prior work on \ac{DER} resolves contact by augmenting the dynamics with a contact energy term \cite{RN773}, whereas our approach explicitly resolves contact using collision detection and a contact solver.

\subsection{Models for Contact}
\Ac{IPC} \cite{RN807} enforces non-penetration using barrier functions. It further employs continuous collision detection within its Newton-based solver to guarantee non-penetration. While highly effective for graphics-oriented simulations, \ac{IPC} can be computationally expensive and may be too slow for robotics applications.

On the other hand, simulating multibody dynamics with frictional contact typically requires solving a \ac{NCP} and can generally only be solved to suboptimal levels in practice \cite{RN860}. To improve computational tractability, convex approximations of the contact problem have been formulated \cite{RN765}. More recent work has introduced regularization to render the problem strictly convex and selected regularization parameters to better match physical contact phenomena \cite{RN1511, RN784}.

\vspace{\baselineskip}
In the following sections, we will overview our methodological framework as follows:
\begin{itemize}[leftmargin=*]
    \item \cref{sec:der-mechanics} describes the \ac{DER} mechanics.
    \item \cref{sec:contact-kinematics} introduces the point and patch contact model.
    \item \cref{sec:dynamics-solver} describes the convex contact solver.
\end{itemize}
While the contact solver is based on prior work, our novelty lies in the correction of the \ac{DER} dynamics, the adaptation of patch contact to slender filaments, and the integration of all components.

\section{Modeling of Filament and Contact}
We model filaments using \ac{DER}s. Both the articulated rigid system and the \ac{DER} are described using generalized coordinates. The full system configuration is denoted by \mbox{$\vec{q} \in \mathbb{R}^{n_q}$}, and the generalized velocity by \mbox{$\vec{v} \in \mathbb{R}^{n_v}$}, where $n_q$ and $n_v$ represent the dimensions of the configuration and velocity vectors, respectively. The time derivative of $\vec{q}$ is related to $\vec{v}$ through a kinematic map \mbox{$\mat{N}(\vec{q}): \mathbb{R}^{n_q} \rightarrow \mathbb{R}^{n_q \times n_v}$}:
\begin{equation}
    \dot{\vec{q}} = \mat{N}(\vec{q})\, \vec{v} .
\end{equation}
For the \ac{DER} \ac{DoFs}, the corresponding submatrix of $\mat{N}(\vec{q})$ is simply the identity. We now detail the kinematics and dynamics specific to the \ac{DER} model.

\subsection{Mechanics of \texorpdfstring{\ac{DER}}{DER}}  \label{sec:der-mechanics}
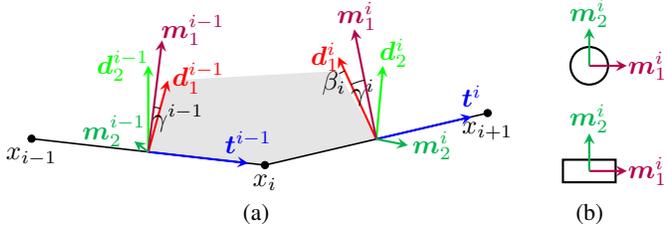
\begin{figure}
    \centering
    \subcaptionbox{\label{fig:der-schematic-a}}{\tdplotsetmaincoords{50}{-20}

\begin{tikzpicture}[tdplot_main_coords, scale=1.5]
  \coordinate (x1) at ($({-2.1*cos(30)}, {2.1*sin(30)}, 0)$);
  \coordinate (x2) at (0,0,0);
  \coordinate (x3) at (2.1,0,0);

  \coordinate (c1) at ($0.5*(x1) + 0.5*(x2)$);
  \coordinate (t1) at ($(c1) + ({0.9*cos(30)}, {-0.9*sin(30)}, 0)$);
  \coordinate (d1_1) at ($(c1) + ({sin(30)}, {cos(30)}, 0)$);
  \coordinate (d2_1) at ($(c1) + (0,0,1)$);
  \coordinate (m1_1) at ($(c1) + ({ sin(30)*cos(45)}, { cos(30)*cos(45)}, {sin(45)})$);
  \coordinate (m2_1) at ($(c1) + ({-sin(30)*sin(45)}, {-cos(30)*sin(45)}, {cos(45)})$);

  \coordinate (c2) at ($0.5*(x2) + 0.5*(x3)$);
  \coordinate (t2) at ($(c2) + (0.9,0,0)$);
  \coordinate (d1_2) at ($(c2) + (0,{ cos(10)},{sin(10)})$);
  \coordinate (d2_2) at ($(c2) + (0,{-sin(10)},{cos(10)})$);
  \coordinate (m1_2) at ($(c2) + (0,{ cos(55)},{sin(55)})$);
  \coordinate (m2_2) at ($(c2) + (0,{-sin(55)},{cos(55)})$);
  \coordinate (dref_2) at ($(c2) + (0,1,0)$);

  \fill[gray, opacity=0.2] (d1_1) -- (c1) -- (x2) -- (c2) -- (dref_2) -- cycle;

  \filldraw (x1) circle[radius=0.9pt] node[below]{$x_{i-1}$};
  \filldraw (x2) circle[radius=0.9pt] node[below]{$x_i$};
  \filldraw (x3) circle[radius=0.9pt] node[below]{$x_{i+1}$};

  \draw[line width=0.6pt] (x1) -- (x2) -- (x3);

  \tikzset{director/.style={thick, ->, >=stealth}}
  \draw[director, blue] (c1) -- (t1) node[above]{$\bm{t}^{i-1}$};
  \draw[director, red] (c1) -- (d1_1) node[right, xshift=-2pt, yshift=1pt]{$\bm{d}_1^{i-1}$};
  \draw[director, green] (c1) -- (d2_1) node[left, xshift=4pt]{$\bm{d}_2^{i-1}$};
  \draw[director, red!70!blue] (c1) -- (m1_1) node[right, xshift=-4pt, yshift=4pt]{$\bm{m}_1^{i-1}$};
  \draw[director, green!70!blue] (c1) -- (m2_1) node[left, xshift=8pt, yshift=4pt]{$\bm{m}_2^{i-1}$};
  \draw pic["$\gamma^{i-1}$", draw=black, angle radius=0.6cm, angle eccentricity=1, pic text options={xshift=8pt, yshift=-4pt}]{angle = d1_1--c1--m1_1};

  \draw[director, blue] (c2) -- (t2) node[above]{$\bm{t}^i$};
  \draw[director, red] (c2) -- (d1_2) node[left, xshift=4pt]{$\bm{d}_1^i$};
  \draw[director, green] (c2) -- (d2_2) node[above, xshift=4pt, yshift=-4pt]{$\bm{d}_2^i$};
  \draw[director, red!70!blue] (c2) -- (m1_2) node[above, xshift=1pt, yshift=-4pt]{$\bm{m}_1^i$};
  \draw[director, green!70!blue] (c2) -- (m2_2) node[right, xshift=-2pt]{$\bm{m}_2^i$};
  \draw pic["$\gamma^i$", draw=black, angle radius=0.8cm, angle eccentricity=1, pic text options={xshift=2pt, yshift=-4pt}]{angle = m1_2--c2--d1_2};
  
  \draw pic["$\beta_i$", draw=black, angle radius=1.0cm, angle eccentricity=1, pic text options={xshift=-3pt, yshift=-4pt}]{angle = d1_2--c2--dref_2};

\end{tikzpicture}

\vspace{-6mm}
\hspace{-2em}} \hfill
    \subcaptionbox{\label{fig:der-schematic-b}}{\begin{tikzpicture}

\def\l{0.5}
\def\d{0.5}
\def\w{0.7}
\def\h{0.3}

\tikzset{director/.style={thick, ->, >=stealth}}

\hspace{3.3mm}

\begin{scope}
    \draw[thick] (0,0) circle(0.5*\d);
    
    \draw[director, red!70!blue]   (0,0) -- (\l,0) node[inner sep=0, right] {$\bm{m}_1^i$};
    \draw[director, green!70!blue] (0,0) -- (0,\l) node[inner sep=0, above] {$\bm{m}_2^i$};
\end{scope}

\begin{scope}[yshift=-1.4cm]
    \draw[thick] (-0.5*\w,-0.5*\h) rectangle (0.5*\w,0.5*\h);
    
    \draw[director, red!70!blue]   (0,0) -- (\l,0) node[inner sep=0, right] {$\bm{m}_1^i$};
    \draw[director, green!70!blue] (0,0) -- (0,\l) node[inner sep=0, above] {$\bm{m}_2^i$};
\end{scope}

\end{tikzpicture}}
    \caption{Schematic of \ac{DER}.
        (a) Centerline of a \ac{DER}. Director $\bm{d}_1^{i-1}$ is rotated about the axis \mbox{$\bm{t}^{i-1} \times \bm{t}^i$} by the angle between $\bm{t}^{i-1}$ and $\bm{t}^i$. The resulting director forms an angle $\beta_i$ with $\bm{d}_1^i$.
        (b) Circular or rectangular cross-section of a \ac{DER}.
    }
    \label{fig:der-schematic}
\end{figure}

The centerline of a \ac{DER} is composed of $n_n$ nodes connected sequentially by \mbox{$n_e = n_n - 1$} edges, as illustrated in \cref{fig:der-schematic}. Each node $i$ is represented by its position \mbox{$\bm{x}_i \in \mathbb{R}^3$} in the world frame. The edge between nodes $i$ and $i+1$ has a unit tangent vector $\bm{t}^i$. We use subscripts to index node-related quantities and superscripts for edge-related ones.
Each edge is associated with a reference frame consisting of directors \mbox{$(\bm{d}_1^i, \bm{d}_2^i, \bm{t}^i)$} and a material frame consisting of directors \mbox{$(\bm{m}_1^i, \bm{m}_2^i, \bm{t}^i)$}. The relative rotation between the two frames is described by the angle \mbox{$\gamma^i \in \mathbb{R}$}.
The material frame rotates with the rod, with $\bm{m}_1^i$ and $\bm{m}_2^i$ aligned with the principal axes of the cross-section (\cref{fig:der-schematic-b}), whereas the reference frame remains fixed in rotation about $\bm{t}^i$.
The configuration vector of the \ac{DER} is given by
\begin{equation}
    \vec{q}_\der = \begin{bmatrix}
        \bm{x}_1^\top & \gamma^1 & \bm{x}_2^\top & \dots & \gamma^{n_e} & \bm{x}_{n_n}^\top
    \end{bmatrix}^\top \in \mathbb{R}^{4 n_n - 1} .
\end{equation}
The subscript $\der$ specifies that the quantities are associated with a \ac{DER}, in contrast to the full system that encompasses both the rigid bodies and all \ac{DER}s.

A \ac{DER} experiences internal elastic forces resulting from stretching, twisting, and bending, under the assumption that shear deformation is negligible. The corresponding internal energy contributions are defined as follows:

\paragraph{Stretching energy}
\begin{equation}
    E_s = \frac{1}{2} \sum_{i=1}^{n_e} E A \left( \frac{\|\bm{e}^i\|}{\|\bar{\bm{e}}^i\|} - 1 \right)^2 \|\bar{\bm{e}}^i\| ,
\end{equation}
where \mbox{$\bm{e}^i = \bm{x}_{i+1} - \bm{x}_i$}, and $\bar{\bm{e}}^i$ is the edge vector in the undeformed configuration, i.e., the vector between nodes $i$ and \mbox{$i+1$} before any deformation. Here, $E$ is the Young's modulus and $A$ is the cross-sectional area.

\paragraph{Twisting energy}
\begin{equation}
    E_t = \frac{1}{2} \sum_{i=2}^{n_n-1} \frac{GJ}{\bar{l}_i} \left( \tau_i - \bar{\tau}_i \right)^2 ,
\end{equation}
where \mbox{$\tau_i = \gamma^i - \gamma^{i-1} + \beta_i$}, and $\beta_i$ is the signed angle between $\bm{d}_1^{i-1}$ and $\bm{d}_1^i$ after parallel transport to a common frame (see \cref{fig:der-schematic-a}). Here, $\bar\tau_i$ denotes the value of $\tau_i$ in the undeformed configuration, and the undeformed Voronoi length is defined as \mbox{$\bar{l}_i = (\|\bar{\bm{e}}^{i-1}\| + \|\bar{\bm{e}}^i\|) / 2$}. The parameters $G$ and $J$ are the shear modulus and the cross-sectional polar moment of inertia, respectively.

\paragraph{Bending energy}
\begin{equation}
    E_b = \frac{1}{2} \sum_{i=1}^{n_n-1} \frac{E I_1}{\bar{l}_i} \left( \kappa_{1,i} - \bar{\kappa}_{1,i} \right)^2 +
          \frac{1}{2} \sum_{i=1}^{n_n-1} \frac{E I_2}{\bar{l}_i} \left( \kappa_{2,i} - \bar{\kappa}_{2,i} \right)^2 ,
\end{equation}
where the discrete curvature vector is defined as
\begin{equation*}
    \bm{\kappa}_i = \frac{2 \bm{t}^{i-1} \times \bm{t}^i}{1 + \bm{t}^{i-1} \cdot \bm{t}^i} ,
\end{equation*}
and its projections define the scalar curvatures:
\begin{equation*}
    \kappa_{1,i} = \bm{\kappa}_i \cdot \frac{\bm{m}_2^{i-1} + \bm{m}_2^i}{2}, \quad
    \kappa_{2,i} = \bm{\kappa}_i \cdot \left(-\frac{\bm{m}_1^{i-1} + \bm{m}_1^i}{2}\right).
\end{equation*}
Here, $\bar{\kappa}_{1,i}$ and $\bar{\kappa}_{2,i}$ are the corresponding scalar curvatures in the undeformed configuration, and $I_1$ and $I_2$ are the second moments of area of the rod’s cross-section about its two principal axes.

With the total internal elastic energy of the \ac{DER} given by \mbox{$E_{\der,\text{int}} = E_s + E_t + E_b$}, the dynamics the of \ac{DER} are governed by
\begin{equation}  \label{eq:der-dynamics}
    \mat{M}_\der \ddot{\vec{q}}_\der = -\frac{\partial E_{\der,\text{int}}}{\partial \vec{q}_\der} - (\alpha \mat{M}_\der + \beta \mat{K}_\der) \, \dot{\vec{q}}_\der + \vec{F}_{\der,\text{ext}} ,
\end{equation}
where $\mat{M}_\der$ is the diagonal lumped mass matrix, \mbox{$\mat{K}_\der = \partial^2 E_\text{int} / \partial \vec{q}_\der^2$} is the stiffness matrix, $\alpha$, $\beta$ are coefficients for the Rayleigh damping model, and $\vec{F}_{\der,\text{ext}}$ includes external forces.

In our implementation, we derived analytical expressions for the Jacobian and Hessian of $E_\text{int}$, which differ from those provided in \cite[\S8.5]{RN1542}. The corrected forms were validated using forward-mode automatic differentiation.
For example, the derivative of the bending energy $E_b$ with respect to the angle $\gamma_i$ is
\begin{equation*}
\begin{split}
    \frac{\partial E_b}{\partial \gamma_i} =
      & \tfrac{E I_1}{\bar{l}_i} \left( \kappa_{1,i} - \bar{\kappa}_{1,i} \right) \bigl( -\tfrac{1}{2} \bm{m}_1^i \cdot \bm{\kappa}_i \bigr) \\
    + & \tfrac{E I_2}{\bar{l}_i} \left( \kappa_{2,i} - \bar{\kappa}_{2,i} \right) \bigl( -\tfrac{1}{2} \bm{m}_2^i \cdot \bm{\kappa}_i \bigr) \\
    + & \tfrac{E I_1}{\bar{l}_{i+1}} \left( \kappa_{1,i+1} - \bar{\kappa}_{1,i+1} \right) \bigl( -\tfrac{1}{2} \bm{m}_1^i \cdot \bm{\kappa}_{i+1} \bigr) \\
    + & \tfrac{E I_2}{\bar{l}_{i+1}} \left( \kappa_{2,i+1} - \bar{\kappa}_{2,i+1} \right) \bigl( -\tfrac{1}{2} \bm{m}_2^i \cdot \bm{\kappa}_{i+1} \bigr) ,
\end{split}
\end{equation*}
rather than zero. Since the Jacobian directly determines the internal elastic force, using the corrected expressions are essential for accurate dynamics.

\subsection{Kinematics of Contact} \label{sec:contact-kinematics}
A simple model for contact is to approximate the intersection between two bodies as occurring at a single point, where both normal and frictional forces are applied. While effective for rigid or stiff bodies, this assumption can be overly simplistic for soft filaments, where deformation occurs not only along the length of the filament but also laterally. In these cases, modeling contact as distributed over patches rather than points yields more physically realistic results. We present the descriptions for both point and patch contact models below.
\begin{figure}
    \centering
    \subcaptionbox{\label{fig:point-contact}}{\includegraphics[width=0.3\linewidth]{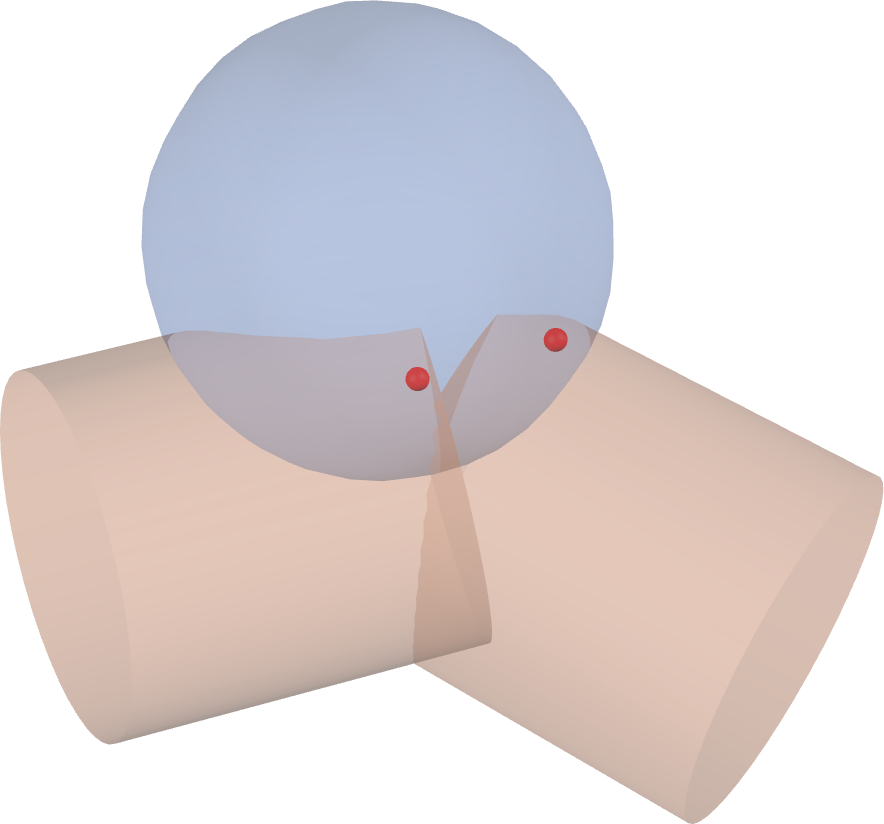}} \hspace{1.5em}
    \subcaptionbox{\label{fig:patch-contact}}{\includegraphics[width=0.3\linewidth]{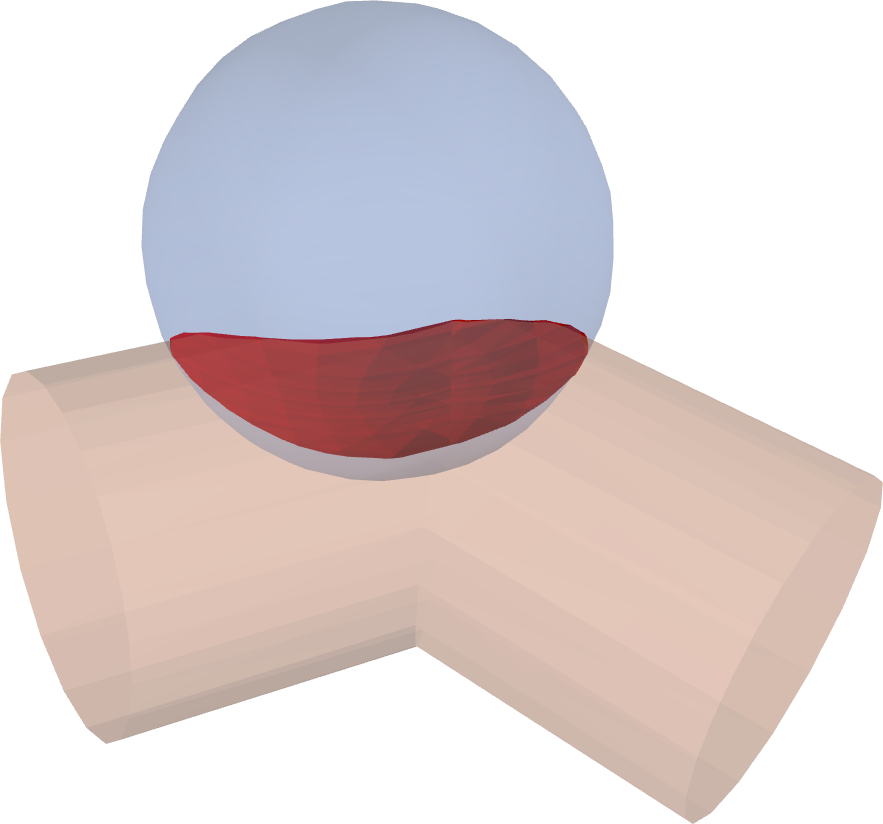}}
    \caption{Contact between a two-segment filament and a ball using different models.
        (a) Point contact, shown as red points.
        (b) Pressure field patch contact, shown as red patches.}
    \label{fig:contact-model}
\end{figure}

\subsubsection{Point Contact}
Each \ac{DER} segment is geometrically approximated as either a cylinder (for circular cross-sections) or a box (for rectangular cross-sections). Other rigid bodies are likewise represented using primitive geometries. A geometry engine then identifies candidate pairs of primitives that are in collision and reports them as contact pairs, excluding neighboring segments of the same filament from collision detection (see \cref{fig:point-contact}).

Each contact pair $i$ is defined by a contact point, the contact normal direction $\bm{n}_i$, and the signed distance $\phi_i$ where \mbox{$\phi_i < 0$} indicates interpenetration. For each contact, we define a local contact frame with the $\bm{n}_i$ direction aligned with the $z$-axis. The relative velocity of contact, \mbox{$\bm{v}_{c,i} \in \mathbb{R}^3$}, is expressed in this frame. It can be decomposed into tangential and normal components as
\begin{equation*}
    \begin{bmatrix} \bm{v}_{c,i} \end{bmatrix} = \begin{bmatrix}
        \bm{v}_{c,i} - v_{n,i} \bm{n}_i \\
        v_{n,i}
    \end{bmatrix}, \quad \text{where } v_{n,i} = \bm{v}_{c,i} \cdot \bm{n}_i.
\end{equation*}

\subsubsection{Pressure Field Patch Contact}
Each \ac{DER} segment is geometrically modeled as a volumetric mesh, with an associated pressure field. This pressure field is defined to be zero on its outer surface and reaches a prescribed maximum along its centerline. Similarly, each rigid body is represented as a volume mesh with a pressure field that is zero on its surface and maximal at its geometric center.
These pressure fields approximate the contact forces that would arise upon interpenetration \cite{RN1513}. According to Newton's third law, the contact patch between two overlapping bodies corresponds to the locus where their pressure fields are equal. A geometry engine detects this contact patch and reports it as a set of polygons (see \cref{fig:patch-contact}).

Each polygon $i$ within the contact patch is characterized by its normal vector $\bm{n}_i$ and area $A_i$, as well as the $p_i$ and pressure gradient $\nabla p_i$ evaluated on its centroid. The stiffness in the direction of the normal is given by
\begin{equation*}
    k = A_i \nabla p_i \cdot (-\bm{n}_i).
\end{equation*}
Since the normal contact force is
\begin{equation*}
    f_n = A_i p_i = -k \phi_i ,
\end{equation*}
the effective signed distance can be expressed as
\begin{equation*}
    \phi_i = \frac{p_i}{\nabla p_i \cdot \bm{n}_i}.
\end{equation*}
Once the effective signed distance $\phi_i$ is computed, the polygon within the contact patch is treated as an equivalent point contact in subsequent calculations.

Assuming there are a total of $n_c$ contact points, stacking the relative velocities of all contact points yields the vector \mbox{$\vec{v}_c \in \mathbb{R}^{3 n_c}$}, which is related to the generalized velocity through the contact Jacobian \mbox{$\mat{J} \in \mathbb{R}^{n_v \times 3n_c}$}:
\begin{equation} \label{eq:contact-jacobian-definition}
    \vec{v}_c = \mat{J}^\top \vec{v} .
\end{equation}

\subsection{Discrete Time Stepping Scheme}
To simulate the system's evolution over time, we divide time into equal intervals of duration $\delta t$, repeatedly advancing the system state from $t_n$ to \mbox{$t_{n+1} = t_n + \delta t$}. To simplify notation, variables evaluated at the current time step $t_n$ are indicated with a subscript $0$, while variables without a subscript refer to values at the next time step $t_{n+1}$.
Following the $\theta$-method formulation in accordance with \cite{RN1537, RN1511}, we define intermediate quantities as
\begin{subequations} \label{eq:theta-method}
\begin{align}
    \vec{q}^\theta &\defeq \theta\, \vec{q} + (1 - \theta)\, \vec{q}_0 , \\
    \vec{v}^\theta &\defeq \theta\, \vec{v} + (1 - \theta)\, \vec{v}_0 , \\
    \vec{v}^{\theta_{vq}} &\defeq \theta_{vq}\, \vec{v} + (1 - \theta_{vq})\, \vec{v}_0 ,
\end{align}
\end{subequations}
which along with
\begin{equation}
    \vec{q} = \vec{q}_0 + \delta t \, \mat{N}(\vec{q}^\theta)\, \vec{v}^{\theta_{vq}} ,
\end{equation}
allows the representation of backwards Euler with \mbox{$\theta = \theta_{vq} = 1$}, symplectic Euler with \mbox{$\theta = 0$}, \mbox{$\theta_{vq} = 1$}, or the midpoint rule with \mbox{$\theta = \theta_{vq} = 1/2$}.

Using \eqref{eq:theta-method}, we formulate the constrained dynamics of the coupled rigid body and \ac{DER} system as
\begin{subequations} \label{eq:constrained-dynamics}
\begin{align}
    &\mat{M}(\vec{q}^\theta) \, (\vec{v} - \vec{v}_0) = \delta t \, \vec{k}(\vec{q}^\theta, \vec{v}^\theta) + \mat{J}(\vec{q}_0)\, \vec{\lambda} , \label{eq:momentum-balance} \\
    &\mathcal{C} \ni \vec{\lambda} \perp \vec{v}_c - \hat{\vec{v}}_c \in \mathcal{C}^* , \label{eq:contact-constraint}
\end{align}
\end{subequations}
where
\begin{equation*}
    \hat{\vec{v}}_c \defeq -\frac{1}{\delta t} \begin{bmatrix}
        0 & 0 & \phi_{1,0} & \dots & 0 & 0 & \phi_{n_c,0}
    \end{bmatrix}^\top \in \mathbb{R}^{3n_c},
\end{equation*}
with $\phi_{i,0}$ the signed distance of the $i$-th contact at the current time step. The convex cone $\mathcal{C}$ is defined as \mbox{$\mathcal{C} \defeq \mathcal{C}_1 \times \dots \times \mathcal{C}_{n_c}$}, where $\mathcal{C}_i$ is the friction cone of the $i$-th contact. The dual cone of $\mathcal{C}$ is written as $\mathcal{C}^*$. Throughout, we use the notation \mbox{$\vec{a} \perp \vec{b}$} to mean \mbox{$\vec{a}^\top \vec{b} = 0$}.

Equation~\eqref{eq:momentum-balance} expresses the momentum balance over the time step, incorporating both rigid body and \ac{DER} \ac{DoFs}. The function \mbox{$\vec{k}(\vec{q}^\theta, \vec{v}^\theta)$} collects all non-contact forces, including Coriolis terms, gravitational forces, and internal elastic forces. The contact impulse vector \mbox{$\vec{\lambda} \in \mathbb{R}^{3n_c}$} captures the net impulses due to contact, including both normal and frictional components.
The contact constraint \eqref{eq:contact-constraint} requires that the contact impulse $\lambda_{n,i}$ and the gap velocity \mbox{$\phi_i / \delta t \approx (\phi_{i,0} + v_{n,i} \delta t) / \delta t$} satisfy a complementarity condition, such that if the contact is separating (\mbox{$\phi_i / \delta t > 0$}), no impulse is applied (\mbox{$\lambda_{n,i} = 0$}); if the contact is maintained (\mbox{$\phi_i = 0$}), a nonzero impulse may be applied.

\section{Dynamics Solver}  \label{sec:dynamics-solver}
This section outlines the procedure used to solve the constrained dynamics in \eqref{eq:constrained-dynamics} to advance the state from one time step to the next. The solver employs a two-stage approach where in the first stage, we compute the unconstrained (i.e., free-motion) velocities that the system would have in the absence of contact. In the second stage, we solve a convex approximation of \eqref{eq:constrained-dynamics} to enforce contact interactions and determine the velocities and contact impulses at the next time step.

We define the momentum residual as
\begin{equation}
    \vec{m}(\vec{v}) = \mat{M}(\vec{q}^\theta) \left(\vec{v} - \vec{v}_0\right) - \delta t \, \vec{k}(\vec{q}^\theta, \vec{v}^\theta).
\end{equation}
In the first stage of the solver, we compute the free-motion velocity $\vec{v}^*$ by solving the nonlinear equation
\begin{equation}  \label{eq:free-motion-velocity}
    \vec{m}(\vec{v}^*) = \vec{0}.
\end{equation}
This equation is solved using the Newton-Raphson method, which iteratively updates the velocity estimate until the residual norm is sufficiently small. Notably, the Hessian of the momentum residual has a fixed sparsity pattern, allowing the reuse of symbolic factorization throughout the iterations.

In the second stage of the solver, we linearize \eqref{eq:constrained-dynamics} around the previously-computed free-motion velocity $\vec{v}^*$. This yields
\begin{subequations} \label{eq:constrained-dynamics-linearized}
\begin{align}
    &\mat{A}\, (\vec{v} - \vec{v}^*) = \mat{J}\, \vec{\lambda} , \label{eq:momentum-balance-linearized} \\
    &\mathcal{C} \ni \vec{\lambda} \perp \vec{v}_c - \hat{\vec{v}}_c \in \mathcal{C}^* . \label{eq:complementarity-constraint}
\end{align}
\end{subequations}
Here, $\mat{A}$ is the matrix obtained by linearizing the momentum balance equation \eqref{eq:momentum-balance} with respect to the generalized velocity $\vec{v}$ about the nominal point \mbox{$\vec{v} = \vec{v}^*$}.
By substituting the \ac{DER} dynamics from \eqref{eq:der-dynamics} into the time-discretized system \eqref{eq:theta-method}--\eqref{eq:constrained-dynamics}, 
the diagonal block of $\mat{A}$ corresponding to the \ac{DER} degrees of freedom is given by
\begin{equation}
    \mat{A}_\der = (1 + \alpha \theta \delta t)\, \mat{M}_\der + \theta \delta t\, (\beta + \theta_{vq} \delta t)\, \mat{K}_\der .
\end{equation}
Because the stiffness matrix $\mat{K}_\der$ may be indefinite, we regularize $\mat{A}_\der$ to ensure positive definiteness.

Since only a subset of the total \ac{DoFs} are involved in contact, the contact Jacobian $\mat{J}$ contains nonzero entries only in the corresponding subset of its rows. To exploit this sparsity, we permute the \ac{DoFs} so that those participating in contact appear first. Under this reordering, \eqref{eq:momentum-balance-linearized} can be written in block form as
\begin{equation} \label{eq:momentum-balance-linearized-blockform}
    \begin{bmatrix}
        \mat{A}_{pp} & \mat{A}_{pn} \\
        \mat{A}_{np} & \mat{A}_{nn}
    \end{bmatrix}
    \begin{bmatrix}
        \vec{v}_p - \vec{v}_p^* \\
        \vec{v}_n - \vec{v}_n^*
    \end{bmatrix}
    =
    \begin{bmatrix}
        \mat{J}_p \, \vec{\lambda} \\
        \vec{0}
    \end{bmatrix} ,
\end{equation}
where the subscript $p$ refers to the participating (contact-related) \ac{DoFs}, and $n$ denotes the non-participating ones.
The Schur complement of $\mat{A}_{pp}$ is given by
\begin{equation*}
    \mat{S}_p = \mat{A}_{pp} - \mat{A}_{pn} \mat{A}_{nn}^{-1} \mat{A}_{np} \succ 0 .
\end{equation*}
Using this, \eqref{eq:momentum-balance-linearized-blockform} can be equivalently expressed as
\begin{subequations}
\begin{align}
    \mat{S}_p \, (\vec{v}_p - \vec{v}_p^*) &= \mat{J}_p \, \vec{\lambda} , \label{eq:momentum-balance-participating} \\
    (\vec{v}_n - \vec{v}_n^*) &= -\mat{A}_{nn}^{-1} \, \mat{A}_{np} \, (\vec{v}_p - \vec{v}_p^*) .
\end{align}
\end{subequations}

The equations \eqref{eq:momentum-balance-participating} and \eqref{eq:complementarity-constraint} are exactly the \ac{KKT} conditions of the following convex optimization problem:
\begin{equation} \label{eq:convex-problem}
\begin{aligned}
    & \minimize_{\vec{v}_p} && \frac{1}{2} \| \vec{v}_p - \vec{v}_p^* \|_{\mat{S}_p}^2 \\
    & \subjectto && \vec{v}_c - \hat{\vec{v}}_c \in \mathcal{C}^* ,
\end{aligned}
\end{equation}
where $\| \vec{x} \|_{\mat{Q}}^2 \defeq \vec{x}^\top \mat{Q} \vec{x}$.
In particular, the Lagrangian of \eqref{eq:convex-problem} is
\begin{equation}
    \mathcal{L}(\vec{v}_p, \vec{\lambda}) = \frac{1}{2} \| \vec{v}_p - \vec{v}_p^* \|_{\mat{S}_p}^2 - \vec{\lambda}^\top (\mat{J}_p^\top \vec{v}_p - \hat{\vec{v}}_c) ,
\end{equation}
where \mbox{$\vec{\lambda} \in \mathcal{C}$} is the dual variable corresponding to the constraint \mbox{$\vec{v}_c - \hat{\vec{v}}_c \in \mathcal{C}^*$}. Recall from \eqref{eq:contact-jacobian-definition} that the stacked contact relative velocity satisfies \mbox{$\vec{v}_c = \mat{J}_p^\top \vec{v}_p$}.

Thus, solving \eqref{eq:convex-problem} directly provides both the primal solution (the participating velocity $\vec{v}_p$) and the dual solution (the contact impulse $\vec{\lambda}$). To solve this problem, we employ the \ac{SAP} solver introduced in \cite{RN1511}, which offers guaranteed convergence for this class of convex formulations.

\section{Results and Discussion}
\begin{table*}
  \centering
  \normalsize
  \caption{Timing and scene statistics.}
  \label{tbl:example-statistics}
  \begin{tabular}{l|l|c|c|c|c}
    \hline
    Example & Contact model & \ac{DoFs} & Time step (s) & Real-time rate & \makecell{Number of contact points\\average / max} \\
    \hline
    Knot tightening             & Point contact & 1205 & 0.0001 & 0.002 &  58 / 164 \\
    Crowned pulley              & Patch contact &  202 & 0.003  & 0.037 & 804 / 917 \\
    Entangling filament gripper & Point contact & 6283 & 0.001  & 0.013 & 104 / 200 \\
    Tying shoelace              & Patch contact &  816 & 0.002  & 0.088 & 164 / 591 \\
    Chain of 100 rings          & Point contact & 8000 & 0.0008 & 0.012 & 396 / 396 \\
    \hline
  \end{tabular}
\end{table*}

\begin{figure}
    \centering
    \begin{tikzpicture}

\definecolor{color1}{HTML}{bebada}
\definecolor{color2}{HTML}{fb8072}
\definecolor{color3}{HTML}{ffffb3}
\definecolor{color4}{HTML}{8dd3c7}

\begin{axis}[
  width=10cm, height=5cm, 
  ybar stacked,                  %
  bar width=12pt,
  ymin=0, ymax=100,              %
  ytick=\empty,
  axis y line=none, 
  symbolic x coords={A,B,C,D,E},     %
  xtick=data,
  xticklabels={%
    \shortstack{Knot\\tightening},
    \shortstack{Crowned\\pulley}, 
    \shortstack{Entangling\\filament\\gripper},
    \shortstack{Tying\\shoelace},
    \shortstack{Chain of\\100 rings}
  },
  legend style={
    at={(0.5,1.05)}, 
    anchor=south, 
    legend columns=2,
    /tikz/every even column/.append style={column sep=1.5em},
    draw=none,
  },
  legend image code/.code={
    \draw[fill=#1,draw=none] (0cm,-0.15cm) rectangle (0.3cm,0.15cm);
  },
]

  \addplot[draw=none, preaction={fill=color1}, pattern=crosshatch,       pattern color=black!80] coordinates {(A, 4.49) (B, 2.39) (C,18.56) (D, 8.02) (E,32.62)};
  \addplot[draw=none, preaction={fill=color2}, pattern=crosshatch dots,  pattern color=black!80] coordinates {(A, 9.73) (B,18.16) (C, 9.53) (D, 5.83) (E, 3.00)};
  \addplot[draw=none, preaction={fill=color3}, pattern=north east lines, pattern color=black!80] coordinates {(A, 1.46) (B, 1.54) (C,12.39) (D, 5.51) (E,15.07)};
  \addplot[draw=none, fill=color4                                                              ] coordinates {(A,84.32) (B,77.91) (C,59.52) (D,80.64) (E,49.31)};
  \legend{\ac{DER} mechanics, Collision detection, Schur complement, \ac{SAP} solver}
  
\end{axis}
  
\end{tikzpicture}
    \caption{Simulation runtime split among computation components.}
    \label{fig:runtime-split}
\end{figure}
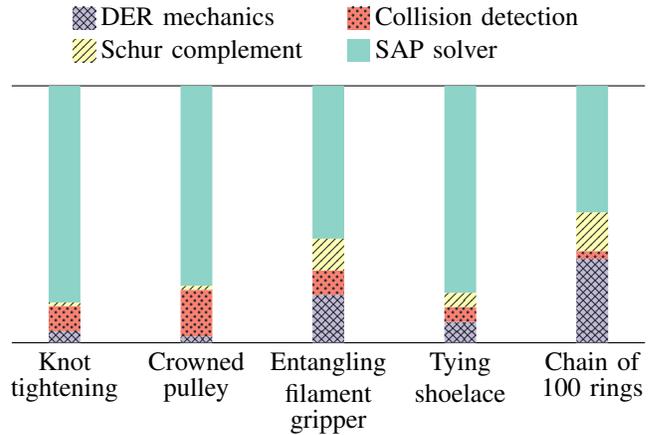
We present a set of test cases to evaluate the accuracy, robustness, and performance of our method, with all simulations executed on a single thread. \glen{maybe just have a separate subsection for the runtime analysis?} \glen{also, can we explain in text why there's variation in the DER mechanics time, e.g., why the chain takes more time? what aspects of each experiment lead to changes in relative runtime?} \cref{tbl:example-statistics} summarizes the contact model and time step size used in each test case, along with the simulation real-time rate (ratio of simulated time to wall-clock time) and the total number of contacts. \cref{fig:runtime-split} further reports the distribution of computation time across the main components of the simulation pipeline. To prevent slender deformable objects from tunneling due to missed collision detections, and to ensure that the linearization in \eqref{eq:constrained-dynamics-linearized} remains a valid approximation, the time step size must be set on the order of \mbox{$1 \times 10^{-3}$~s} or smaller. We also find that patch contact models significantly increase the number of effective contact points, since each contact patch consists of multiple polygons, each treated as an individual contact point. This leads to an increase in the contact resolution workload. Finally, the \ac{SAP} solver consistently dominates runtime, accounting for at least 50\% of the total cost across all test cases, making it the primary computational bottleneck.

\subsection{Capstan Effect}
\begin{figure}
    \centering
    \begin{tikzpicture}

\definecolor{color1}{HTML}{e41a1c}
\definecolor{color2}{HTML}{377eb8}
\definecolor{color3}{HTML}{4daf4a}

\begin{axis}[
    width=8.6cm, height=6cm,
    xlabel={Wrap angle $\varphi$},
    ylabel=$\log(T_2/T_1)$,
    xmin=1.25663706, xmax=6.28318531,
    xtick={1.25663706, 2.51327412, 3.76991118, 5.02654825, 6.28318531},
    xticklabels={$0.4\pi$, $0.8\pi$, $1.2\pi$, $1.6\pi$, $2.0\pi$},    
    ylabel shift=-2mm,
    grid=major,
    legend pos=south east
]

\addplot [black, thick, domain=0:6.3] {0.2*x};
\addlegendentry{Theoretical}

\pgfplotstableread{
x y
1.25663706 0.21790192
1.57079633 0.30989827
1.88495559 0.38487413
2.19911486 0.38853504
2.51327412 0.45930646
2.82743339 0.53921486
3.14159265 0.63434655
3.45575192 0.69448822
3.76991118 0.70494359
4.08407045 0.77586489
4.39822972 0.85004823
4.71238898 0.92255764
5.02654825 0.93273155
5.34070751 0.96074257
5.65486678 1.01874796
5.96902604 1.08712440 
6.28318531 1.13422032
}\dataA

\addplot+[only marks, color1, mark=o] table {\dataA};
\addlegendentry{$\Delta\varphi = \pi/10$}
\addplot [color1, thick, forget plot, dashed] table [
    y={create col/linear regression={y=y}}
] {\dataA};

\pgfplotstableread{
x y
1.25663706 0.23936111
1.57079633 0.29528664
1.88495559 0.38130768
2.19911486 0.41202924
2.51327412 0.50125795
2.82743339 0.55611133
3.14159265 0.61274358
3.45575192 0.69901799
3.76991118 0.73015248
4.08407045 0.81836399
4.39822972 0.87249501
4.71238898 0.92793558
5.02654825 1.01233561
5.34070751 1.04281110
5.65486678 1.12841118
5.96902604 1.18105125
6.28318531 1.23453308
}\dataB

\addplot+[only marks, color2, mark=square] table {\dataB};
\addlegendentry{$\Delta\varphi = \pi/20$}
\addplot [color2, thick, dashed, forget plot] table [
    y={create col/linear regression={y=y}}
] {\dataB};

\pgfplotstableread{
x y
1.25663706 0.24664914
1.57079633 0.31183020
1.88495559 0.37719777
2.19911486 0.44002706
2.51327412 0.50366000
2.82743339 0.56912434
3.14159265 0.63540746
3.45575192 0.70098627
3.76991118 0.76312293
4.08407045 0.82598818
4.39822972 0.89016000
4.71238898 0.95466150
5.02654825 1.01796187
5.34070751 1.07730838
5.65486678 1.13789978
5.96902604 1.19955310
6.28318531 1.26133841
}\dataC

\addplot+[only marks, color3, mark=x] table {\dataC};
\addlegendentry{$\Delta\varphi = \pi/40$}
\addplot [color3, thick, forget plot, dashed] table [
    y={create col/linear regression={y=y}}
] {\dataC};

\end{axis}

\begin{scope}[shift={(1.8,3.6)}]
    \def\R{0.5}
    \def\L{\R*1.4}
    \def\ang{135}
    
    \draw[fill=gray!20] (0,0) circle (\R);
    
    \coordinate (A) at (\R,0); %
    \coordinate (B) at ({\R*cos(\ang)},{\R*sin(\ang)}); %
    \coordinate (O) at (0,0); %

    \draw[line width=1.5pt] (\R,-\L) node[below, inner sep=2pt]{$T_1$} -- (A);
    \draw[line width=1.5pt] (A) arc[start angle=0, end angle=\ang, radius=\R];
    \draw[line width=1.5pt] (B) -- ({\R*cos(\ang) - \L*sin(\ang)},{\R*sin(\ang) + \L*cos(\ang)})
      node[below left, inner sep=0pt]{$T_2$};

    \draw[dashed, dash pattern=on 2pt off 1pt] (O) -- (A);
    \draw[dashed, dash pattern=on 2pt off 1pt] (O) -- (B);
    \node at (0,0) [anchor=south west, inner sep=0pt, yshift=2pt] {$\varphi$};
\end{scope}

\end{tikzpicture}
    \caption{Ratio of pulling forces at both ends of a rope as a function of wrap angle around a cylindrical post. Each rope segment has length $R\,\Delta\varphi$, where $R$ is the radius of the cylindrical post.}
    \label{fig:capstan-effect}
\end{figure}
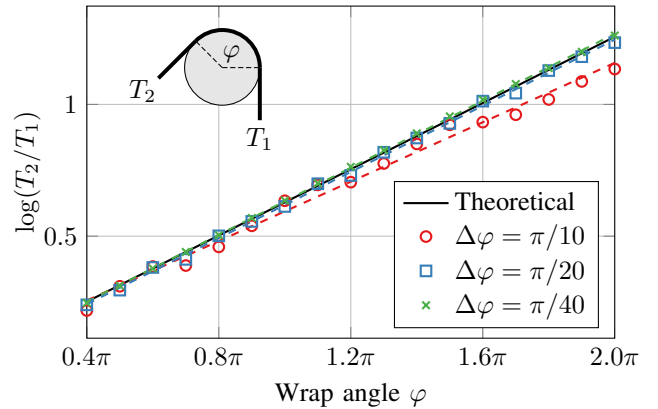
We simulate the capstan effect, in which a rope is wrapped around a cylindrical post through a wrap angle $\varphi$ and subjected to pulling forces $T_1$ and $T_2$ at its two ends. Due to frictional interaction between the rope and the capstan surface, the ratio of transmitted forces grows exponentially with the wrap angle. The maximum pulling force that can be sustained is given by the classical expression $T_2 = T_1 e^{\mu \varphi}$, where $\mu$ is the coefficient of friction \cite[\S6.8]{meriamStatics2015}.

In our setup, the dynamic coefficient of friction $\mu$ between the cylindrical post and the rope is set to 0.2, and self-collision of the rope is disabled to isolate interation between the rope and the cylindrical post. One end of the rope is controlled via a \ac{PD} controller to track a fixed position, while the other end is gradually pulled with increasing force. Once steady state is reached, the ratio of forces at both ends is recorded, and the experiment is repeated for multiple wrap angles $\varphi$. The results, shown in \cref{fig:capstan-effect}, indicate that the simulation accurately reproduces the expected exponential force amplification. A mesh refinement study further confirms that finer discretizations, where each rope segment spans an angle $\Delta\varphi$ around the post, converge toward the theoretical prediction, validating the accuracy of our frictional contact model.

\subsection{Knot Tightening}
We simulate the tightening of a knot with an unknotting number of four (\cref{fig:knot-tightening}), corresponding to a configuration where the rope braids around itself four times.
In the simulation, the rope is pulled at both ends with a constant velocity of 0.8 m/s, while the applied force is regulated by a \ac{PD} controller. Due to the high pulling speed, the time step is set to \mbox{$1 \times 10^{-4}$ s}. The result of our framework (\cref{fig:knot-tightening-b}) closely matches that of a real rope (\cref{fig:knot-tightening-a}), and successfully reproduces the phenomenon of snap-through buckling: once the knot becomes sufficiently tight, it abruptly transitions from a near-circular shape to a collapsed configuration.

In contrast, the method in \cite{RN773}, which augments the dynamics with a contact energy term to enforce non-penetration, fails under the same material properties, pulling speed, and time step size (\cref{fig:knot-tightening-c}). While the contact energy dominates when the rope is loose, it becomes comparatively small relative to the internal elastic energy as the rope tightens, resulting in tunneling.

For comparison, we also model the rope as a chain of rigid capsules with virtual springs to emulate bending and twisting stiffness (\cref{fig:knot-tightening-d}). Although this model avoids tunneling, it fails to reproduce the characteristic snap-through buckling, instead producing a non-physical tightened configuration.

\subsection{Crowned Pulley}
\begin{figure}
    \centering
    \subcaptionbox{\label{fig:crowned-pulley-a}}{
        \includegraphics[height=0.12\linewidth]{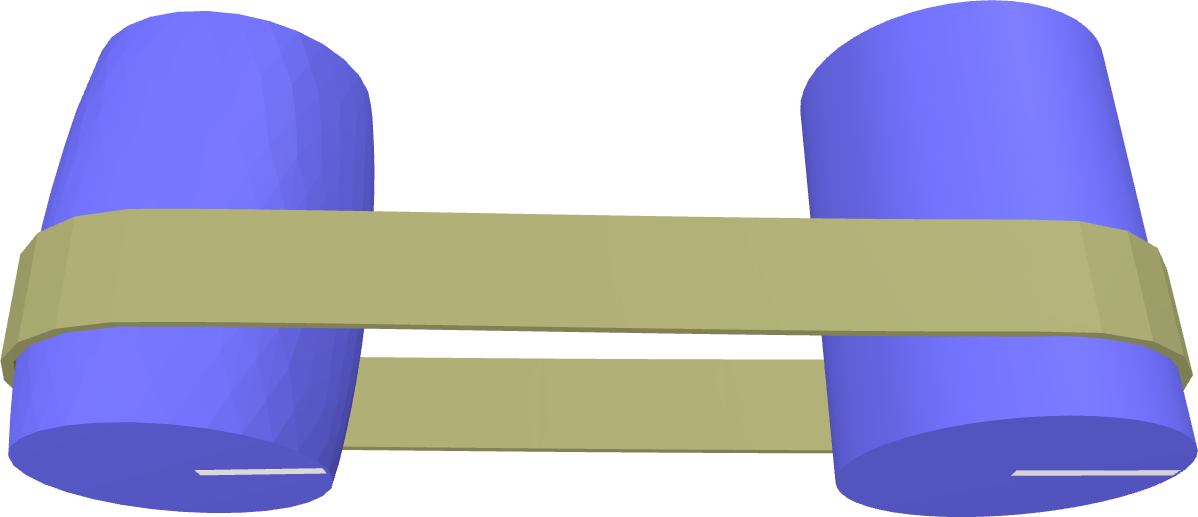} \hspace{3mm}
        \includegraphics[height=0.12\linewidth]{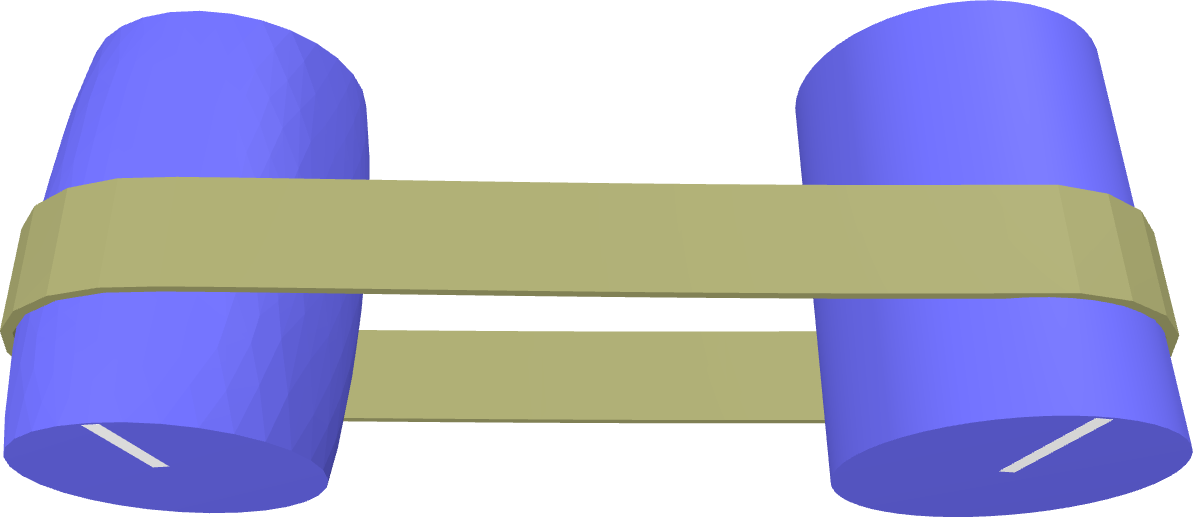} \hspace{3mm}
        \includegraphics[height=0.12\linewidth]{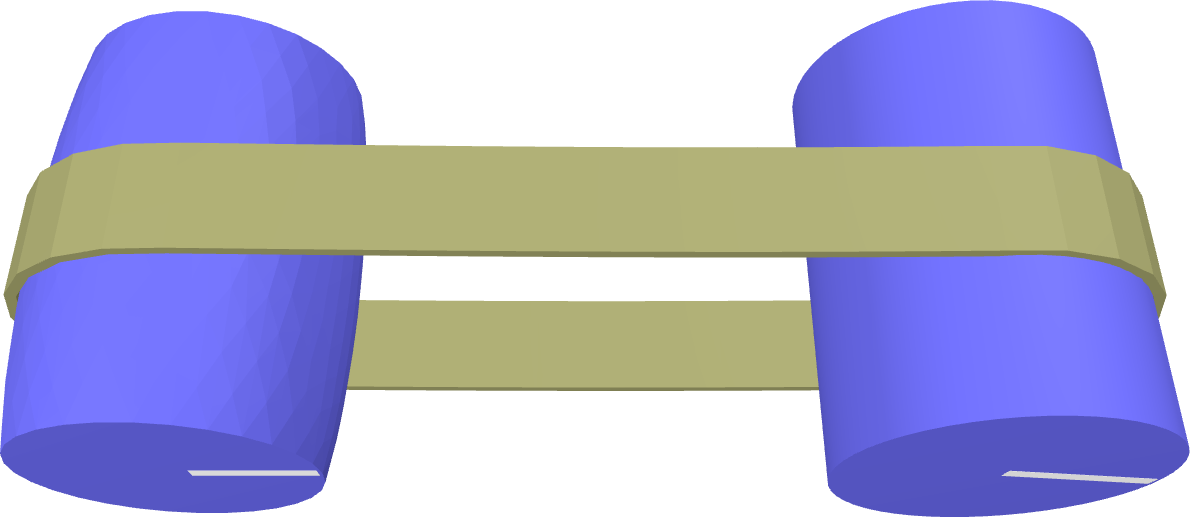}
    }
    \subcaptionbox{\label{fig:crowned-pulley-b}}{
        \includegraphics[height=0.12\linewidth]{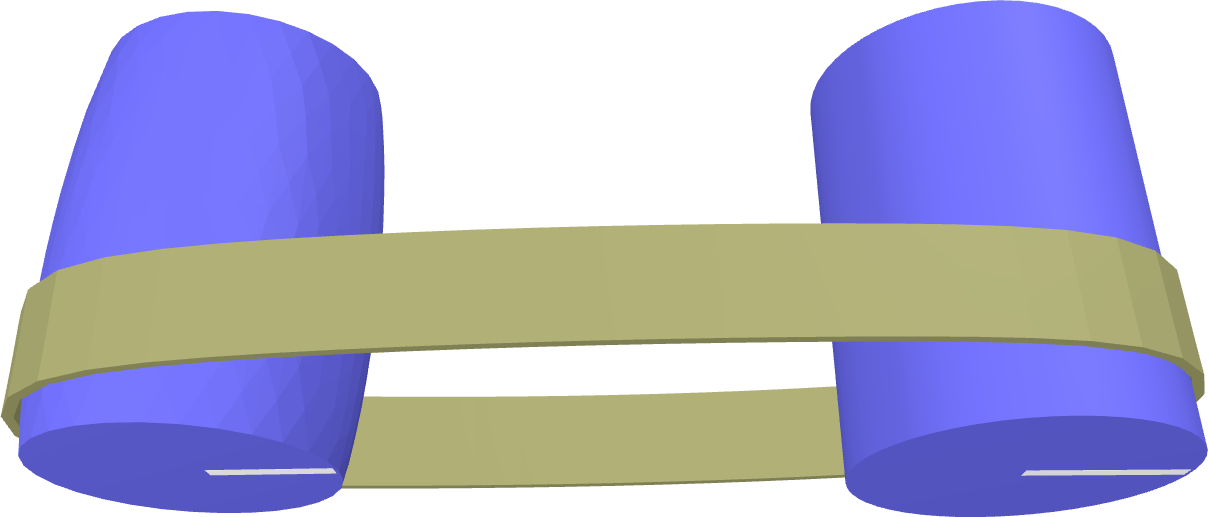} \hspace{3mm}
        \includegraphics[height=0.12\linewidth]{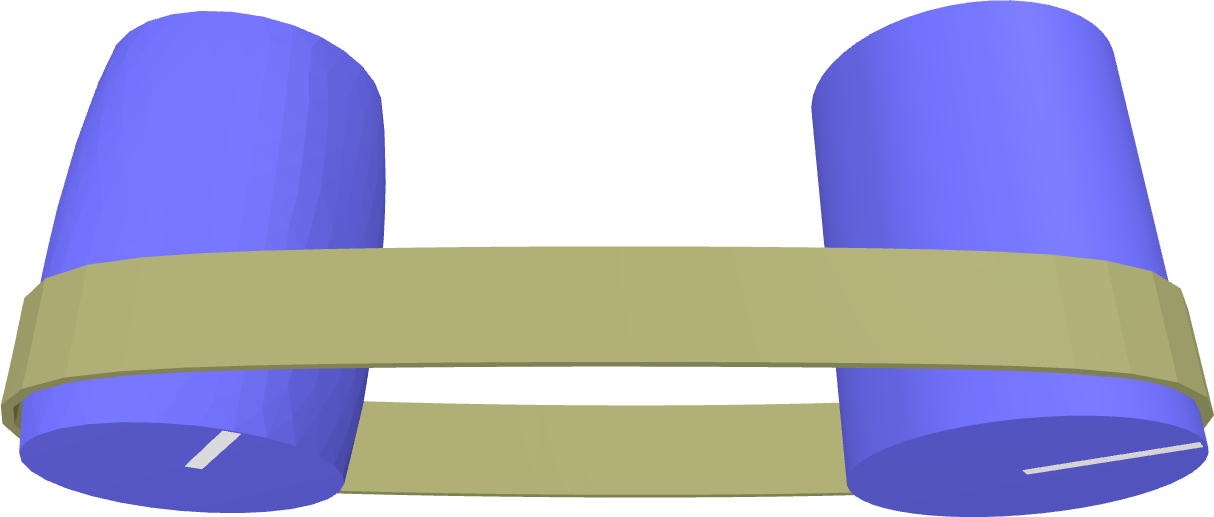} \hspace{3mm}
        \includegraphics[height=0.12\linewidth]{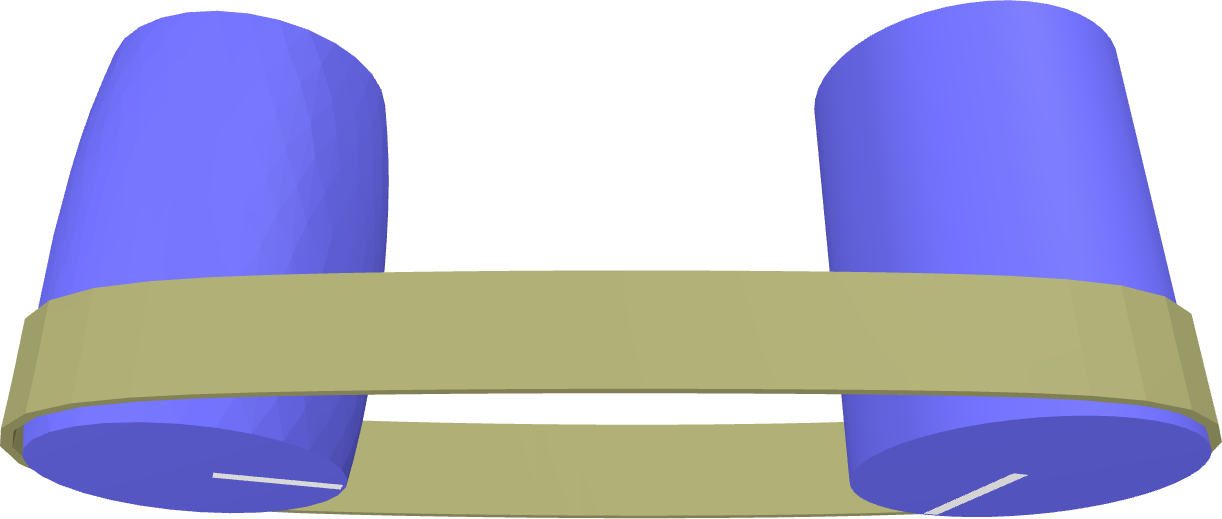}
    }
    \caption{Keyframes of a belt running over a pair of pulleys, with the left pulley having a crowned profile. (a) Our framework simulates the self-centering of the belt as the pulley rotates. (b) Volumetric \ac{FEM} \cite{RN1537} fails to capture even the belt tensioning.}
    \label{fig:crowned-pulley}
\end{figure}
A crowned pulley has a convex surface, with a slightly larger diameter at the center that tapers toward the edges. When a flat belt runs over such a pulley, it naturally self-centers \cite[\S17.1]{Shigley2011}.
We simulate a belt riding on a crowned pulley, discretized so that each segment has a length equal to one-twentieth of the pulley’s circumference (\cref{fig:crowned-pulley-a}). The belt configuration is represented by the nodal positions $\bm{x}_i$ of its centerline together with twisting angles $\gamma^i$, and is assigned a wide rectangular cross-section. Although this representation does not capture bending across the belt’s width, it is sufficient to reproduce the self-centering phenomenon.

This behavior emerges only when both the distributed contact between the belt and pulley and the belt’s internal stretching and twisting are modeled. If any component is omitted, the self-centering effect disappears: a point contact model cannot generate the distributed forces needed to support the belt on the pulley’s curved surface, neglecting stretching prevents proper belt tensioning, and neglecting twisting prevents the segments from forming the geometry that drives self-centering.

We compare against the volumetric \ac{FEM} in \cite{RN1537} (\cref{fig:crowned-pulley-b}). Even with a mesh resolution in the thickness direction equal to one-third of the belt thickness, the volumetric \ac{FEM} fails to capture belt tensioning and, as a result, cannot form the geometry required for self-centering. This failure is likely due to the large element aspect ratio: the mesh is coarse along the belt’s longitudinal and width directions but fine in the thickness direction, which artificially increases bending stiffness and prevents the belt from deforming correctly. Using an even finer mesh to reduce the aspect ratio would be computationally prohibitive.\glen{do we use a finer mesh to simulate our filament? if so, say something like: in contrast, we are able to handle finer meshes while remaining computationally efficient because ...}

\subsection{Entangling Filament Gripper}
\begin{figure}
    \centering
    \adjincludegraphics[width=0.32\linewidth, trim={{.3\width} 0 {.3\width} 0}, clip]{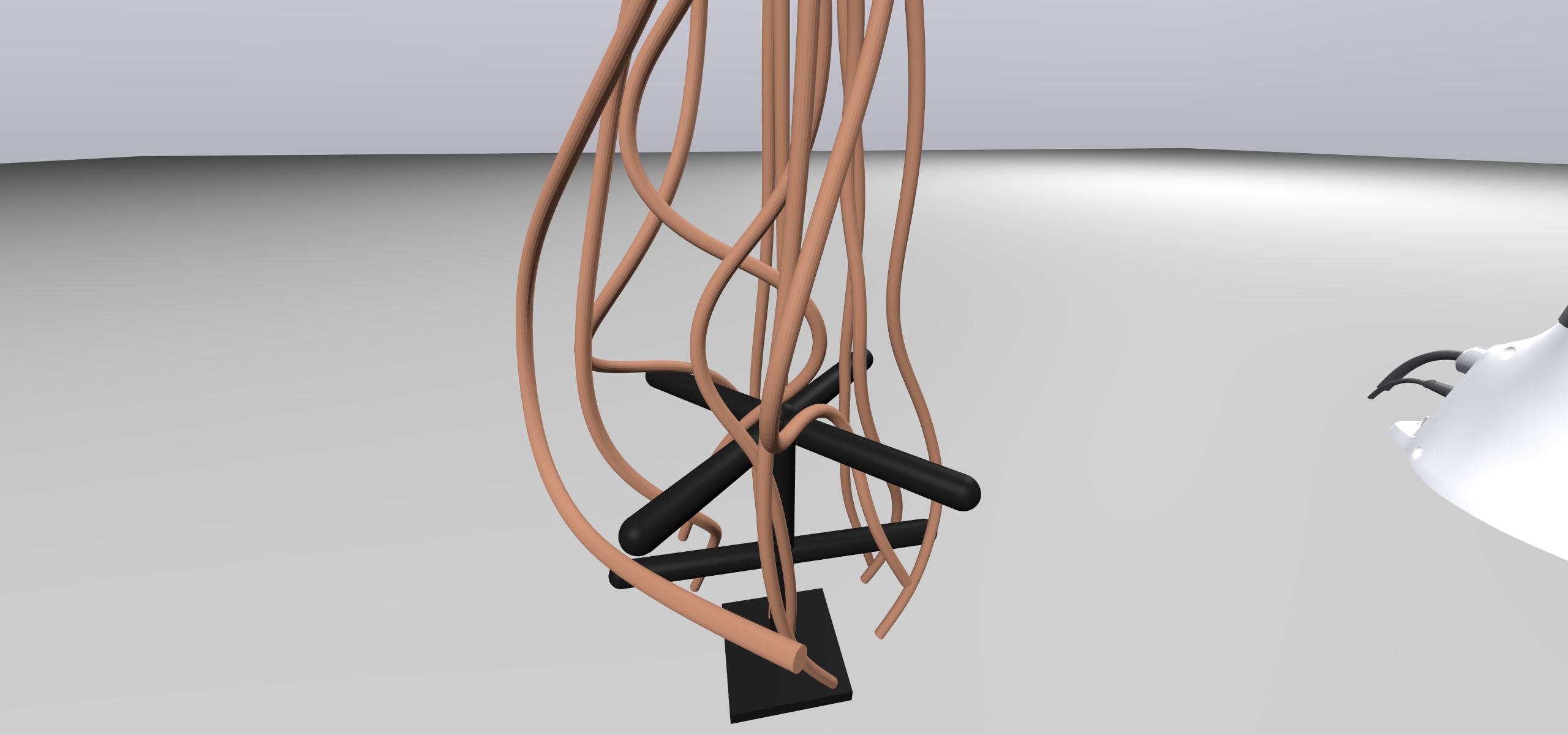} \hfill
    \adjincludegraphics[width=0.32\linewidth, trim={{.3\width} 0 {.3\width} 0}, clip]{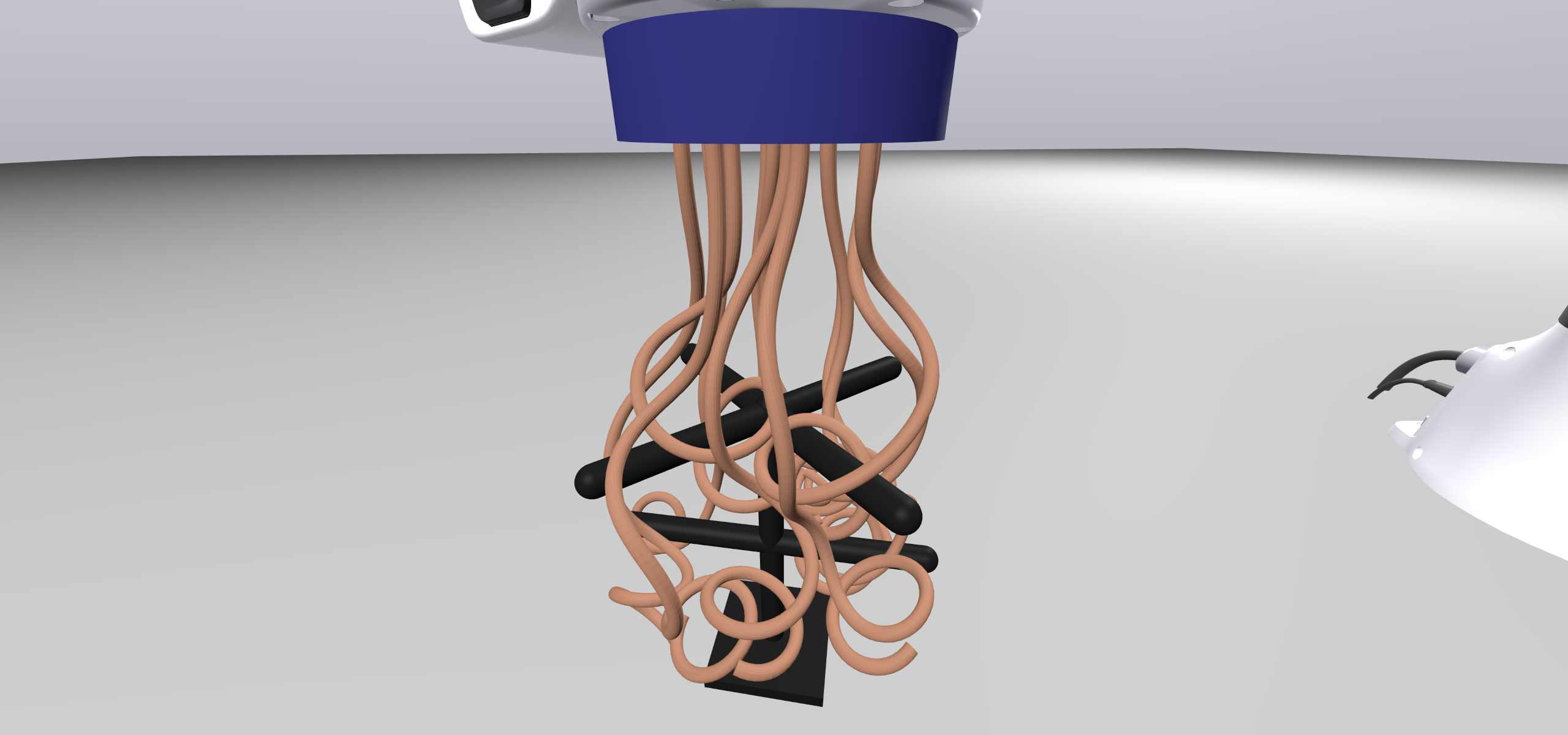} \hfill
    \adjincludegraphics[width=0.32\linewidth, trim={{.3\width} 0 {.3\width} 0}, clip]{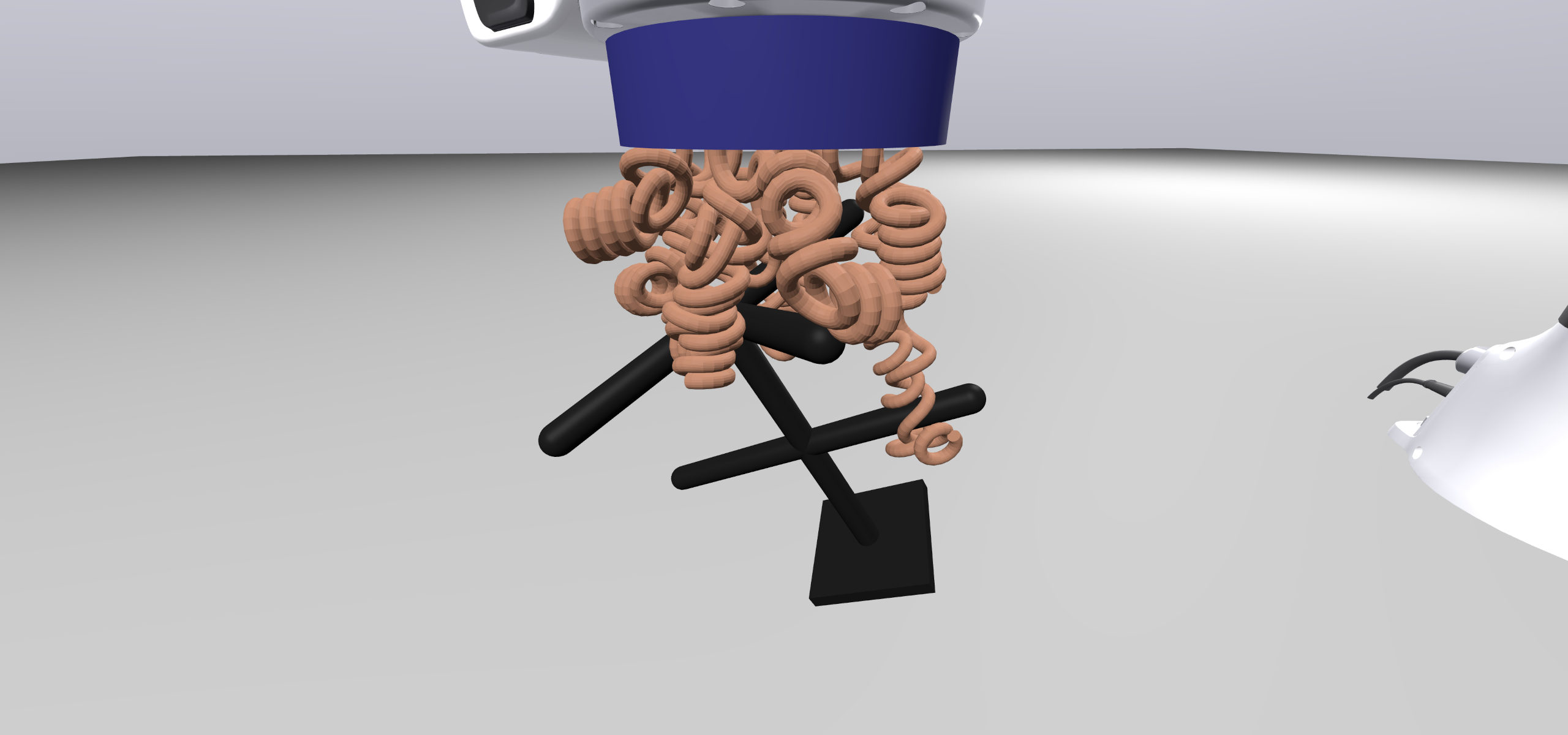}
    \caption{Keyframes of the entangling filament gripper grasping a branched object. The gripper initially approaches the object from above, and as the filaments’ natural curvature increases, it achieves a stochastic grasp on the object.}
    \label{fig:entangling-filament-gripper}
\end{figure}
Another application of filament–rigid body interaction simulation lies in the study of novel soft robotic grippers \cite{RN832, RN833}. In particular, we simulate an entangling filament gripper \cite{RN832} grasping a branched object as illustrated in \cref{fig:entangling-filament-gripper}. This type of gripper is composed of multiple slender elastomer filaments whose natural curvature can be actuated through pneumatic pressure. Grasping is achieved by exploiting the stochastic self-entanglement and cross-entanglement of the filaments, enabling the gripper to grasp topologically-complex objects. In our simulations, the directions of natural curvature are held fixed (the ratio $\bar{\kappa}_{1i} / \bar{\kappa}_{2i}$ remains constant), while the overall curvature is gradually increased (simultaneously increasing $\bar{\kappa}_{1i}$ and $\bar{\kappa}_{2i}$) to achieve grasping. The simulation robustly resolves contact interactions among filaments, between filaments and the branched object, and between filaments and the gripper base. Furthermore, owing to the use of implicit time integration, our simulations can advance with significantly larger time steps than the \mbox{$1 \times 10^{-5}$~s} step size used in the original study \cite{RN832}, which employs Elastica \cite{RN819} as the simulator.\glen{do you have a comparison where you run Elastica at $10^{-3}$ step size and it fails to accurately simulate? also, how do we know Fig 7 is accurate? did you compare with the Elastica paper?}

\subsection{Tying Shoelace}
\begin{figure}
    \centering
    \subcaptionbox{\label{fig:tying-shoelace-a}}{
        \adjincludegraphics[width=0.32\linewidth, trim={{.25\width} 0 {.25\width} 0}, clip]{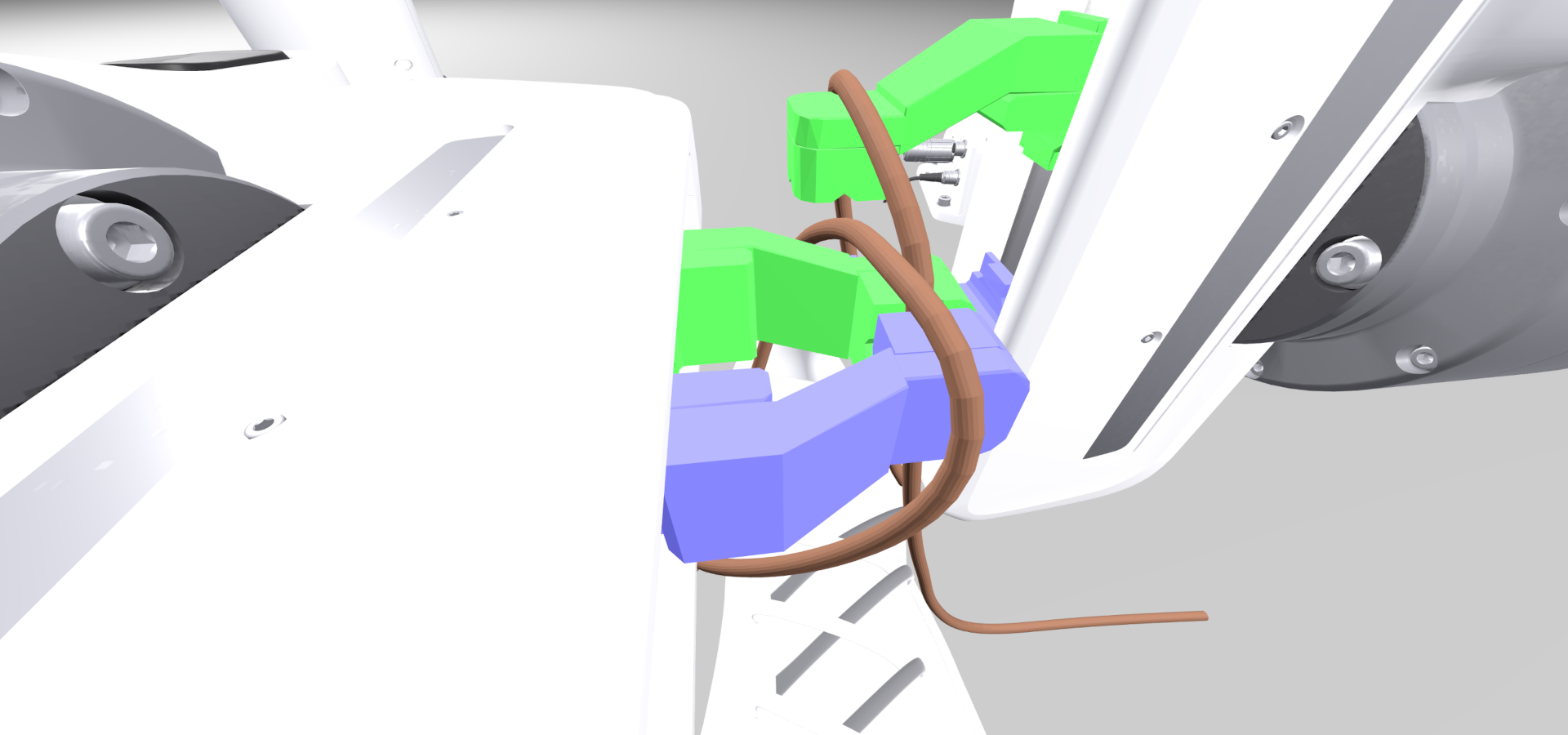} \hspace{-0.7em}
        \adjincludegraphics[width=0.32\linewidth, trim={{.25\width} 0 {.25\width} 0}, clip]{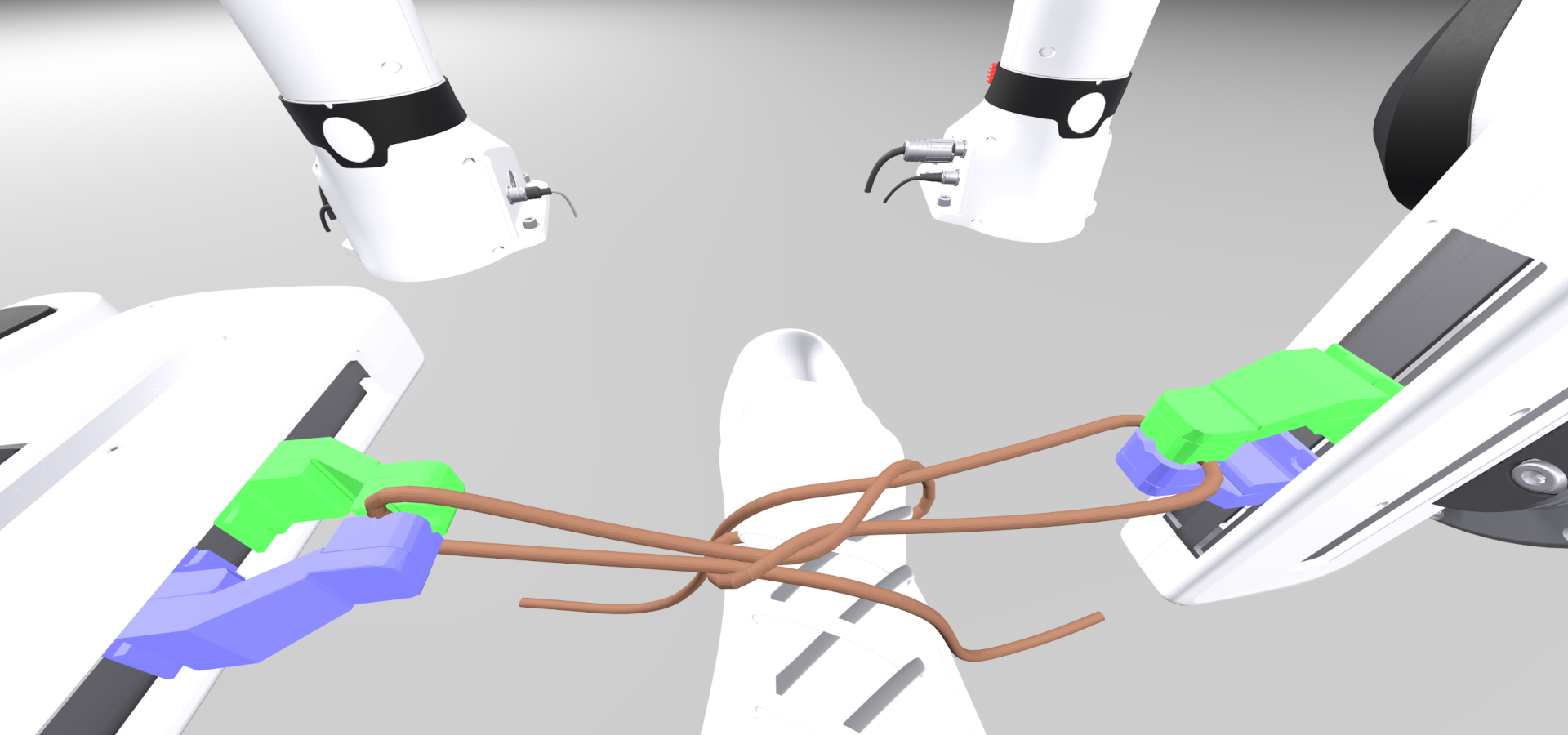}
    } \hspace{-1.2em}
    \subcaptionbox{\label{fig:tying-shoelace-b}}{
        \adjincludegraphics[width=0.32\linewidth, trim={{.25\width} 0 {.25\width} 0}, clip]{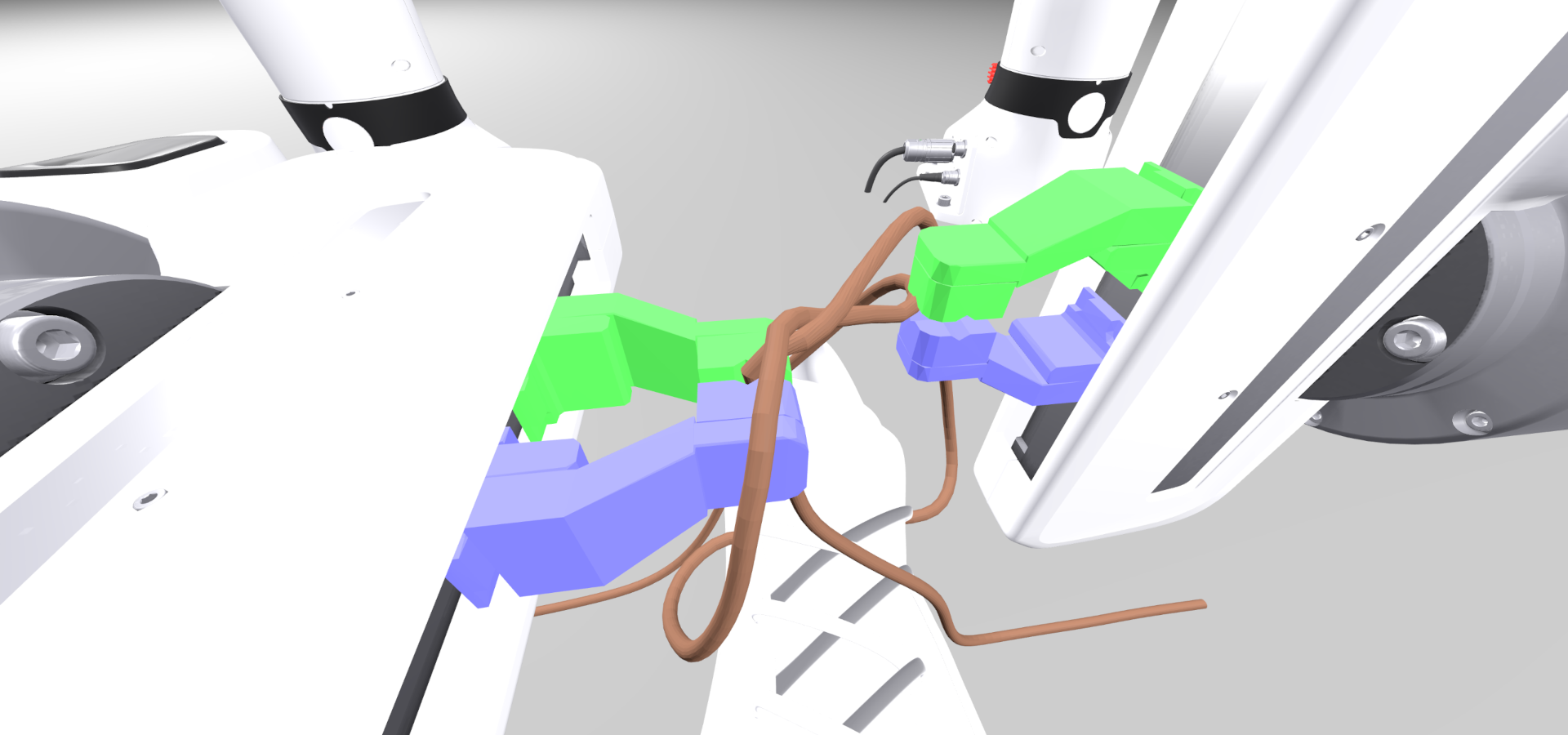}
    } \newline
    \raisebox{3mm}{
    \subcaptionbox{\label{fig:tying-shoelace-c}}{
        \adjincludegraphics[width=0.32\linewidth, trim={{.25\width} 0 {.25\width} 0}, clip]{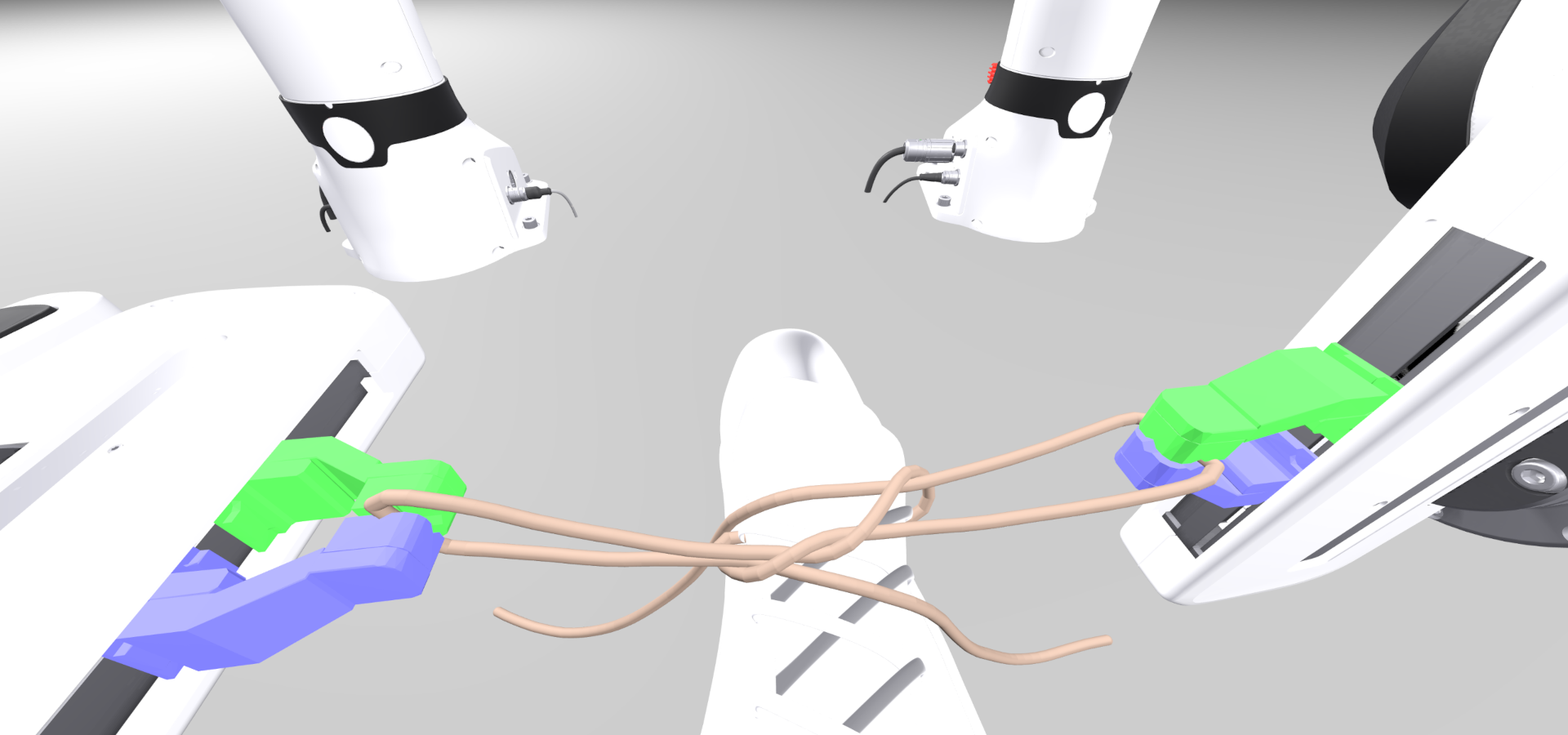}
    }} \hfill
    \subcaptionbox{\label{fig:tying-shoelace-d}}{
        \input{tying-shoelace_pinching-force}
    }
    \subcaptionbox{\label{fig:tying-shoelace-e}}{
        \input{tying-shoelace_elastic-energy}
    }
    \caption{Bimanual robotic arms tying a shoelace.
        (a) \textbf{\ac{DER} with patch contact} successfully simulates the full shoelace-tying sequence.
        (b) \textbf{\ac{DER} with point contact} fails, as the shoelace slips out of the fingers.
        (c) \textbf{Capsule-chain-and-springs with patch contact} produces visually convincing results.
        (d) Pinching forces of the gripper fingers.
        (e) Internal elastic energy of the shoelace with different models.
    }
    \label{fig:tying-shoelace}
\end{figure}
To further demonstrate frictional and compliant contact between filaments and rigid bodies, we simulate bimanual robot arms tying a shoelace. The arms follow a predefined open-loop trajectory. Using the pressure field patch contact model for both the shoelace and gripper fingers successfully simulates the full shoelace tying sequence (\cref{fig:tying-shoelace-a}).

In contrast, a point contact model causes the shoelace to slip from the gripper fingers, preventing task completion (\cref{fig:tying-shoelace-b}). The difference is evident in the finger pinching forces (\cref{fig:tying-shoelace-d}). With patch contact, the force smoothly increases until it balances the \ac{PD}-controlled finger forces, producing stable friction. With point contact, however, the pinching force fluctuates violently, then drops as the shoelace escapes and the fingers close completely. These oscillations stem from how contact is resolved: the solver enforces non-penetration constraints \eqref{eq:contact-constraint} at the end of each step, while the free-motion update \eqref{eq:free-motion-velocity} is applied in the next. The penetration depth after the free-motion update can vary across time steps, leading to unstable force estimates and ultimately preventing steady friction.

For comparison, we also model the shoelace with a chain of rigid capsules with virtual springs to emulate bending and twisting stiffness. By extensively tuning the stiffness and damping parameters, the capsule-chain-and-springs model can produce visually similar results to the \ac{DER} model (\cref{fig:tying-shoelace-c}). However, due to the high stiffness of the springs, this simulation requires a time step of \mbox{$1 \times 10^{-4}$~s}, much smaller compared to the \ac{DER} model. Furthermore, examining the internal elastic energy of both models (\cref{fig:tying-shoelace-e}) reveals that the capsule-chain-and-springs model exhibits much higher frequency fluctuations. These high-frequency fluctuations are an artifact, as such high-frequency content would naturally dissipate quickly in real-world materials.

\subsection{Chain of Rings}
\begin{figure}
    \centering
    \subcaptionbox{\label{fig:ring-chain-a}}{
        \includegraphics[height=0.4\linewidth]{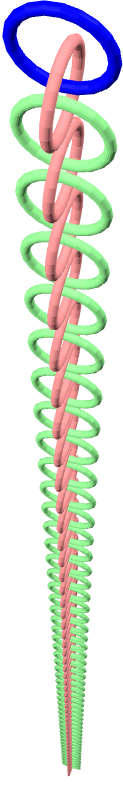}
        \includegraphics[height=0.4\linewidth]{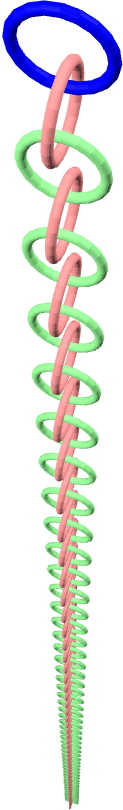}
    } \hfill
    \subcaptionbox{\label{fig:ring-chain-b}}{\begin{tikzpicture}

\definecolor{color1}{HTML}{377eb8}

\begin{axis}[
    width=6.5cm, height=4cm,
    xlabel={Number of rings},
    ylabel={Normalized runtime},
    xmin=0, xmax=100,
    ymin=0, ymax=1,
    xlabel shift=-1mm,  
    ylabel shift=-1mm,
    ylabel style={xshift=-5pt},
    grid=major,
    legend pos=south east
]

\addplot [line width=0.8pt, mark=triangle, color=black]
  table[]{%
5   0.08093606
10  0.11486509
15  0.15966039
20  0.23656354
25  0.34326719
30  0.34545677
35  0.34656884
40  0.38957017
45  0.42160128
50  0.43117895
55  0.47779161
60  0.51656274
65  0.63257331
70  0.68783006
75  0.69868085
80  0.73666691
85  0.77814933
90  0.83267363
95  0.91616580
100 1.00000000
};

\end{axis}

\end{tikzpicture}}
    \caption{A chain of rings suspended under gravity. (a) Keyframes of a 100-ring chain. (b) Runtime for varying number of rings.}
    \label{fig:ring-chain}
\end{figure}
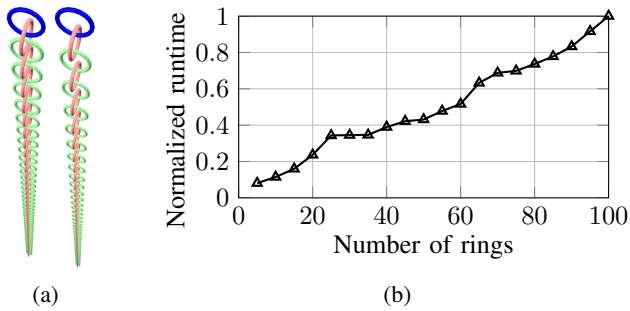
A chain composed of identical rings is simulated under gravity (\cref{fig:ring-chain-a}). Each ring has a radius of 1~cm, a circular cross-section of diameter 2.5~mm, a material density of 500~kg/m³, and a Young’s modulus of 10~MPa. The geometry of each ring is discretized into 20 segments. A time step of \mbox{$8 \times 10^{-4}$~s} is used, small enough to prevent missed collision detections and the resulting tunneling artifacts. To evaluate long-term stability, the simulation was continued for an additional 10~minutes of physical time after the chain had settled under gravity. The chain remained in static equilibrium throughout this period, confirming the stability of the simulation over extended durations.

To assess the scalability of our framework, we simulate chains with varying numbers of rings and measure the runtime (\cref{fig:ring-chain-b}). The runtime exhibits linear scaling with the number of rings. This behavior is largely due to the example exhibiting stable contact patterns, which allows the \ac{SAP} solver to reuse the symbolic factorization in its internal Cholesky decomposition.

\section{Conclusion}
We presented a convex formulation of frictional contact between filaments and rigid bodies, combining physically accurate \ac{DER} dynamics and contact mechanics. Contact is modeled using either point contact, suited for efficiency and approximately fixed cross-sections, or pressure field patch contact, which captures lateral deformation and fine-grained friction. Both models are stateless, and contact is resolved by partitioning out non-participating \ac{DoFs} and solving with the \ac{SAP} solver, which guarantees global optimality.

Through a series of test cases, we validated the accuracy of frictional forces and the robustness of our framework in handling scenes with dense contacts. We also compared our approach against baseline methods, demonstrating improvements in physical fidelity in enforcing non-penetration, producing stable frictional forces, and maintaining smooth internal elastic energy.

Several limitations point to promising directions for future work. Profiling indicates that the \ac{SAP} solver accounts for the majority of computational cost, motivating exploration of parallelization strategies, particularly for the Cholesky decomposition. Additionally, the requirement for small time steps remains a bottleneck for simulation speed. Adaptive time-stepping strategies could allow larger time steps in scenarios with sparse contacts.

% Generated by IEEEtran.bst, version: 1.14 (2015/08/26)

\end{document}